\documentclass[compsoc,conference,a4paper,10pt,times]{IEEEtran}
\IEEEoverridecommandlockouts
\usepackage{cite}
\usepackage{amsmath,amssymb,amsfonts}
\usepackage{algorithmic}
\usepackage{graphicx}
\usepackage{textcomp}
\usepackage{bmpsize}
\usepackage{xcolor}
\usepackage{lipsum}
\usepackage[colorlinks=true,urlcolor=black]{hyperref}
\def\BibTeX{{\rm B\kern-.05em{\sc i\kern-.025em b}\kern-.08em
    T\kern-.1667em\lower.7ex\hbox{E}\kern-.125emX}}

\usepackage{slashbox}
\usepackage{multirow}
\usepackage{comment}
\usepackage{amsthm}
\usepackage{bm}

\usepackage{amsmath,amssymb}
\usepackage{graphicx}
\usepackage{xspace}
\usepackage{xcolor}
\usepackage{stmaryrd}
\usepackage{subcaption}
\usepackage{array}
\usepackage{hyperref}
\usepackage{tablefootnote}
 
\newcommand{\descr}[1]{\vspace{0.1cm}\noindent\textit{#1}}
\usepackage[utf8]{inputenc}%(only for the pdftex engine)
\newtheorem{definition}{Definition}
\newtheorem{theorem}{Theorem}

\usepackage{stmaryrd}
\usepackage[ruled,noline, linesnumbered]{algorithm2e} 
\usepackage{slashbox}
\usepackage{makecell}

\SetAlgoNoLine
\SetAlCapNameFnt{\small}
\SetAlCapFnt{\small}

\newcommand{\mbf}[1]{{\mathbf{#1}}}

\newcommand{\Tcl}{T_{\mathsf{cl}}}
\newcommand{\Tgd}{T_{\mathsf{gd}}}

\newcommand{\TP}{\textit{TP }}
\newcommand{\TN}{\textit{TN }}
\newcommand{\TPR}{\textit{TPR }}
\newcommand{\TNR}{\textit{TNR }}
\newcommand{\Pos}{\textit{P }}
\newcommand{\Neg}{\textit{N }}
\newcommand{\FPR}{\textit{FPR }}

\newcommand{\AUC}{\textit{AUROC }}   

\makeatletter
\newcommand{\linebreakand}{%
  \end{@IEEEauthorhalign}
  \hfill\mbox{}\par
  \mbox{}\hfill\begin{@IEEEauthorhalign}
}
\makeatother
    
\begin{document}

\title{Compression Boosts Differentially Private Federated Learning}%\\ %{\footnotesize
    %\textsuperscript{*}Note: Sub-titles are not captured in Xplore and
    
    %\thanks{A preprint of this paper has been
    %deposited on ArXiv.}  }

\author{\IEEEauthorblockN{Raouf Kerkouche}
\IEEEauthorblockA{\textit{Privatics team} \\
\textit{Univ. Grenoble Alpes, Inria}\\
38000 Grenoble, France \\
raouf.kerkouche@inria.fr}
\and
\IEEEauthorblockN{Gergely \'Acs}
\IEEEauthorblockA{\textit{Crysys Lab} \\
\textit{BME-HIT}\\
Budapest, Hungary \\
acs@crysys.hu}
\and
\IEEEauthorblockN{Claude Castelluccia}
\IEEEauthorblockA{\textit{Privatics team} \\
\textit{Univ. Grenoble Alpes, Inria}\\
38000 Grenoble, France \\
claude.castelluccia@inria.fr}
%\and
\linebreakand 
\IEEEauthorblockN{Pierre Genev\`es}
\IEEEauthorblockA{\textit{Tyrex team} \\
\textit{Univ. Grenoble Alpes, CNRS,} \\
\textit{Inria, Grenoble INP, LIG}\\
38000 Grenoble, France \\
pierre.geneves@cnrs.fr}
%\and
%\IEEEauthorblockA{\textit{dept. name of organization (of %Aff.)} \\
%\and
%\IEEEauthorblockN{6\textsuperscript{th} Given Name %Surname}
%\IEEEauthorblockA{\textit{dept. name of organization (of %Aff.)} \\
}

\maketitle

\begin{abstract}

Federated  Learning  allows distributed entities to train a common model  collaboratively without sharing their own data. Although it prevents data collection and aggregation by exchanging only parameter updates, it remains vulnerable to various inference and reconstruction attacks where a malicious entity can learn private information about the participants' training data from the captured gradients.
Differential Privacy is used to obtain theoretically sound privacy guarantees against such inference attacks by noising the exchanged update vectors. However, the added noise is proportional to the model size which can be very large with modern neural networks. This can result in poor model quality.
In this paper, compressive sensing is used to reduce the model size and hence increase model quality without sacrificing privacy. 
We show experimentally, using 2 datasets, that our privacy-preserving proposal can reduce the communication costs by up to 95\% with only a negligible performance penalty compared to traditional non-private federated learning schemes.

%Our proposal outperforms standard differentially private federated learning depending on the dataset. 

%Besides boosting differential 
%the size of the model is often equal to million of parameters and then the noise will become too large, resulting in a very inaccurate updates model. Therefore, reducing the model size become indispensable if we aim to get a good accuracy with differential privacy. To this purpose, we have decided to use a scalable compressive sensing algorithm for compression. We present a non-private scheme based on compressive sensing and an extension which use in addition differential privacy. Surprisingly, the compressive sensing is based on an optimization problem which seems adapted to a scenario where noise is added as with Differential Privacy. Our results show that it is possible to be bandwidth efficient for slight accuracies degradation. 

\end{abstract}


\section{Introduction}
\label{sec:intro}

%\begin{enumerate}
%    \item Federetad Learning (few sentences). WHy is it important, useful?
%    \item Two problems: compression and privacy. WHy are they challenging (compression: secure aggregation, privacy: gradients leak lot of information see DLG attack and Reza's attacks)
%    \item Solutions so far: quantization, (sketching?), DP. Instead, we leverage sparsity to compress model updates.

%    \item Our solution: CS + DP. Our contirbution is twofold: First, we use CS for compression. Second, we show that CS boosts DP federated learning.
%\end{enumerate}

%Our specific contributions are as follows:
%\begin{itemize}
%    \item compression improves DP-SGD in federated learning (btw, would that help in centralized learning??)
%    \item we propose a CS variant that is scalable and has good accuracy. our approach select the first-k instead of random k (we don't have streaming)
%    \item CS is superior to other approaches 
%    \item  CS can be combined with secure aggregation
%\end{itemize}

Traditional training of machine learning models usually requires the centralization of the user-held data. This limitation is considerably penalizing especially when the data is sensitive such as medical data. To deal with this problem, federated learning protocols have been proposed\cite{ShokriS15,FedAVG} to collaboratively train  a common model without sharing any private training data held by individual parties.
In federated learning, each entity trains a common model using its own training data, and share only the gradients (i.e., model update) with each other through a central server. The server updates the common model with the shared gradients, and re-distributes the updated model to the clients for further training. This process repeats until the  convergence of the common model.

%a server sent periodically a global model to a set of entities, each entity trains the model before to send it back to the server, the server averages the received models to obtain an updated one. 
However, sharing gradients computed by individual parties can leak information about their private training data. Several recent attacks have demonstrated that a sufficiently skilled adversary, who can capture the model updates (gradients) sent by individual parties, can infer whether a specific record \cite{NasrSH19} or a group property \cite{Property_inference} is present in the dataset of a specific party. Moreover, complete training samples can also be reconstructed purely from the captured gradients \cite{ZhuLH19}.

Differential privacy (DP) \cite{Dwork2014book} has become a de facto privacy model which provides a formal privacy guarantee for any participant. It guarantees that the common model is roughly independent of any single client's training data, and depends only on the characteristics that are shared among multiple parties' training data\footnote{If client-level DP is considered. See Section \ref{sec:privacy_model} for more clarification.}.    
DP can be achieved by adding Gaussian noise to the shared model updates. In addition, secure aggregation protocols \cite{BonawitzIKMMPRS16} allow parties to add noise to the model update in a distributed manner, which  increases robustness against byzantine attacks and requires less noise than other decentralized perturbation approaches such as randomized response \cite{ErlingssonPK14} used in local differential privacy \cite{Hybrid-approach}. 

%gradient or model's update which aims to protect for a record-level protection or a client-level protection, respectively. 

However, the norm of the added noise is proportional to the model size (i.e., the number of model parameters or weights). Indeed, the noise is added to every coordinate value of the gradient (update) vector including those which have very small magnitude and would anyway not improve convergence. In other words, adding noise to sparse model updates can slow down convergence significantly, or result in poor model quality \cite{zhu2020votingbased}. 

In this paper, we propose to first lossily compress the gradients using compressive sensing \cite{donoho2006compressed,candes2006near,candes2006robust} and then add noise to the compressed gradient vector. The noisy compressed vectors are then transferred to the server for aggregation. This approach has several benefits. First, compressed gradients are less sparse and also shorter than the original gradient vectors. This allows to add less noise to the compressed gradients, which eventually yields faster convergence  with more accurate models than the uncompressed noisy gradients. Second, compressive sensing is linear, which means that the sum of compressed gradients equals the compressed sum of the gradients. Therefore, compressive sensing can be smoothly integrated with secure aggregation;
the server can only access the aggregated compressed vectors which is identical to the compressed aggregation.
Finally, by decreasing the size of the model updates, communication costs are reduced and bandwidth is saved. This is crucial with resource constrained parties training large models which is not uncommon nowadays.

The main contributions of this paper are summarized as follows:
\begin{itemize}
    \item We use a slighlty modified version of compressive sensing to compress sparse model updates in federated learning.  Our protocol, called FL-CS allows to save bandwidth and reduce communication costs by transferring only the low frequency components of the gradient vector to the server (instead of some random frequency components like in traditional compressive sensing). The server can reconstruct the approximated sparse gradient vector by efficiently solving a convex quadratic optimization problem. 
    This approach provides more accurate reconstruction than simply applying the inverse Fourier transform on the low frequency components.
    Our approach is scalable to large gradient vectors and is almost as accurate as the vanilla federated learning protocol, referred to FL-STD, without any compression, still incuring much smaller communication cost.  
    \item We propose a privacy-preserving extension of FL-CS, called FL-CS-DP, by adding Gaussian noise to the compressed gradients. In FL-CS-DP, participants inject Gaussian noise in a distributed manner so that the sum of the noisy compressed vectors is differentially private. In addition, secure aggregation guarantees that the server (or any other third party) can only learn the noisy compressed aggregate owing to the linear compression scheme.  
    Reconstructing the approximated gradients is an instance of Basis Pursuit Denoising (or LASSO), which can be solved with efficient solvers that provide large accuracy despite the added Gaussian noise.
    We show that FL-CS-DP produces more accurate models than FL-STD-DP, that is, the differentially private variant of the vanilla federated learning protocol without any compression. Therefore, compression boosts the accuracy of differentially private federated learning and also reduces bandwith cost by more than 60\% with early stopping \cite{KERAS_early_stopping}. 
    \item We evaluate our proposals on real datasets, a private medical dataset of 1.2 millions of US hospital patients and the public Fashion-MNIST dataset.
    %We show that FL-CS-DP has an accuracy improvement of 15\% compared to  FL-STD-DP, while it reduces its bandwidth cost with more than 60\%.
    %Moreover, the larger the compression ratio the more accurate FL-CS-DP is.
    We show that FL-CS-DP reduces its bandwidth cost with more than 60\% compared to FL-STD-DP, meanwhile suffering negligible performance loss  compared  to  uncompressed  federated  learning  without any  privacy  guarantee (FL-STD).
\end{itemize}

\section{Background}
\label{sec:backg}
\subsection{Federated Learning (FL-STD)}
\label{FL-STANDARD}

In federated learning \cite{ShokriS15,FedAVG}, multiple parties (clients) build a common machine learning model from union of their training data without sharing them with each other. At each round of the training, a selected set of clients retrieve the global model from the parameter server, update the global model based on their own training data, and send back their updated model to the server. The server aggregates the updated models of all clients to obtain a global model that is re-distributed to some selected parties in the next round.  

In particular, a subset $\mathbb{K}$ of all $N$ clients are randomly selected at each round to update the global model, and $C = |\mathbb{K}| / N$ denotes the fraction of selected clients. At round $t$, a selected client $k \in \mathbb{K}$ executes $\Tgd$ local gradient descent iterations on the common model $\mbf{w}_{t-1}$ using its own training data $D_k$ ($D = \cup_{k\in \mathbb{K}} D_k$), and obtains the updated model $\mbf{w}_{t}^k$, where the number of weights is denoted by $n$ (i.e., $|\mbf{w}_t^{k}| = |\Delta \mbf{w}_t^k| = n$ for all $k$ and $t$). Each client $k$ submits the update $\Delta \mbf{w}_{t}^k = \mbf{w}_{t}^k - \mbf{w}_{t-1}^k$ to the server, which then updates the common model as follows: $\mbf{w}_{t} = \mbf{w}_{t-1} + \sum_{k \in \mathbb{K}} \frac{|D_k|}{\sum_j |D_j|} \Delta \mbf{w}_{t}^k$, where $|D_k|$ is known to the server for all $k$ (a client's update is weighted with the size of its training data).
The server stops training after a fixed number of rounds $\Tcl$, or when the performance of the common model does not improve on a held-out data. 

Note that each $D_k$ may be generated from different distributions (i.e., Non-IID case), that is, any client's local dataset may not be representative of the population distribution \cite{FedAVG}. This can happen, for example, when not all output classes are represented in every client's training data. 
The federated learning of neural networks is summarized in Alg.~\ref{alg:fed_learn}. In the sequel, each client is assumed to use the same model architecture.

\begin{algorithm}[h]
\small
		\caption{FL-STD: Federated Learning \label{alg:fed_learn}}
	\DontPrintSemicolon
	{\bf Server:}\;
	\Indp Initialize common model $w_0$\;
	\For {$t=1$ \KwTo $\Tcl$}
	{
	    Select $\mathbb{K}$ clients uniformly at random \;
		\For {\textrm{each} client $k$ \textrm{in} $\mathbb{K}$}
		{	
			$\Delta \mbf{w}_t^k = \mathbf{Client}_k(\mbf{w}_{t-1})$\;
		}
		$\mbf{w}_{t} = \mbf{w}_{t-1} + \sum_{k} \frac{|D_k|}{\sum_j^N |D_j|} \Delta \mbf{w}_{t}^{k}$\;
	}
	\KwOut{Global model $\mbf{w}_t$}\;
	\Indm {\bf $\mathbf{Client}_{k}(\mbf{w}_{t-1}^k)$:}\;
	\Indp
	$\mbf{w}_{t}^k = \mathbf{SGD}(D_k, \mbf{w}_{t-1}^k, \Tgd)$\;
	\KwOut{Model update $(\mbf{w}_{t}^k- \mbf{w}_{t-1}^k)$} 
\end{algorithm}

\begin{algorithm}[h]
\small
	\caption{Stochastic Gradient Descent \label{alg:sgd}}
	\DontPrintSemicolon
	\KwIn{$D$ : training data, $\Tgd$ : local epochs, $\mathbf{w}$ : weights}  
    \For {$t=1$ \KwTo $\Tgd$}
	{
	    Select batch $\mathbb{B}$ from $D$ randomly\;
	    $\mbf{w} = \mbf{w} - \eta \nabla f(\mathbb{B}; \mbf{w})$\;
	}
    \KwOut{Model $\mbf{w}$} 
\end{algorithm}

%\begin{itemize}
%    \item Non-IID: When any user's local dataset is not representative of the population distribution \cite{FedAVG}. For example, if the learning task is to classify 10 digits, each user will have only examples of some digits. 
%    \item IID: When any user's local dataset is representative of the population distribution. 
%\end{itemize}

The motivation of federated learning is three-fold: first, it aims to provide confidentiality of each participant's training data by sharing only model updates instead of potentially sensitive training data. Second, in order to decrease communication costs, clients can perform multiple local SGD iterations before sending their update back to the server. 
Third, in each round, only a few clients are required to perform local training of the common model, which further diminishes communication costs  and makes the approach especially appealing with large number of clients.

%Indeed, in other collaborative learning techniques \cite{}, \emph{all} clients return their model updates to the server after processing a single  mini-batch, which can result in large communication overhead when the number of clients is large. 
However, several prior works have demonstrated that model updates do leak potentially sensitive information \cite{NasrSH19,Property_inference}. Hence, simply not sharing training data \emph{per se} is not enough to guarantee their confidentiality.

\subsection{Differential Privacy}
\label{sec:DP}
Differential privacy allows a party to privately release information about a dataset:  a function of an input dataset is perturbed, so that any information which can differentiate a record from the rest of the dataset is bounded~\cite{Dwork2014book}.
%Consequently, for a record owner, it means that any privacy breach will not be due to participating in the database. 

 \begin{definition}[Privacy loss]
 Let $\mathcal{A}$ be a privacy mechanism which assigns a value $\mathit{Range}(\mathcal{A})$ to a dataset $D$. The privacy loss of $\mathcal{A}$ with datasets $D$ and $D'$ at output $O \in \mathit{Range}(\mathcal{A})$ is a random variable $\mathcal{P}(\mathcal{A},D,D',O) = \log\frac{\Pr[\mathcal{A}(D) = O]}{\Pr[\mathcal{A}(D') = O]}$ 
 %\vspace{-0.2cm}
 %$$
 %\mathcal{P}(\mathcal{A},D,D',O) = \log\frac{\Pr[\mathcal{A}(D) = O]}{\Pr[\mathcal{A}(D') = O]} \vspace{-0.2cm}
 %$$
 where the probability is taken on the randomness of $\mathcal{A}$.%\vspace{-0.25cm}
 \label{def:ploss}
 \end{definition}

\begin{definition}[$(\epsilon,\delta)$-Differential Privacy~\cite{Dwork2014book}] 
A privacy mechanism $\mathcal{A}$ guarantees $(\varepsilon, \delta)$-differential privacy if for any database $D$ and $D'$, differing on at most one record, $\Pr_{O \sim \mathcal{A}(D)}[\mathcal{P}(\mathcal{A},D,D',O) > \varepsilon] \leq \delta$. 

\label{def:DP}
\end{definition}

Intuitively, this guarantees that an adversary, provided with the output of $\mathcal{A}$, can draw almost the same conclusions (up to $\varepsilon$ with probability larger than $1 - \delta$) about any record no matter if it is included in the input of $\mathcal{A}$ or not~\cite{Dwork2014book}. That is, for any record owner, a privacy breach is unlikely to be due to its participation in the dataset.

\descr{Moments Accountant.} Differential privacy maintains composition; the privacy guarantee of the  $k$-fold adaptive composition  of $\mathcal{A}_{1:k} = \mathcal{A}_1, \ldots, \mathcal{A}_k$ can be computed using the moments accountant method \cite{Abadi}. In particular, it follows from Markov's inequality that $\Pr[\mathcal{P}(\mathcal{A},D,D',O) \geq \varepsilon] \leq \mathbb{E}[\exp(\lambda \mathcal{P}(\mathcal{A},D,D',O))]/\exp(\lambda\varepsilon)$ for any output $O \in \mathit{Range}(\mathcal{A})$ and $\lambda > 0$. This implies that $\mathcal{A}$ is $(\varepsilon, \delta)$-DP %(see Definition~\ref{def:DP}) 
with $\delta = \min_{\lambda} \exp(\alpha_{\mathcal{A}}(\lambda) - \lambda \varepsilon)$, where $\alpha_{\mathcal{A}}(\lambda) = \max_{D,D'} \log\mathbb{E}_{O\sim \mathcal{A}(D)}[\exp(\lambda \mathcal{P}(\mathcal{A},D,D',O))]$ is the log of the moment generating function of the privacy loss. The privacy guarantee of the composite mechanism $\mathcal{A}_{1:k}$ can be computed using that $\alpha_{\mathcal{A}_{1:k}}(\lambda) \leq \sum_{i=1}^k \alpha_{\mathcal{A}_{i}}(\lambda)$ \cite{Abadi}. \smallskip 

\descr{Gaussian Mechanism.} There are a few ways to achieve DP, including the Gaussian mechanism~\cite{Dwork2014book}. A fundamental concept of all of them is the \emph{global sensitivity} of a function~\cite{Dwork2014book}.
%\vspace{-0.1cm}
\begin{definition}[Global $L_p$-sensitivity] 
For any function $f:\mathcal{D} \rightarrow \mathbb{R}^ n$, the $L_p$-sensitivity of $f$ is
$\Delta_p f = \max_{D, D'} || f(D)-f(D') ||_p$, 
for all $D, D'$ differing in at most one record, where $||\cdot||_p$ denotes the $L_p$-norm.\vspace*{-0.15cm}
\label{def:global_sens}
\end{definition}
\smallskip
The Gaussian Mechanism~\cite{Dwork2014book} 
consists of adding Gaussian noise to the true output of a function.
In particular, for any function $f:\mathcal{D} \rightarrow \mathbb{R}^n$, the Gaussian mechanism is defined as adding i.i.d Gaussian noise with variance $(\Delta_2 f \cdot \sigma)^2$  and zero mean to each coordinate value of  $f(D)$. Recall that the pdf of the Gaussian distribution with mean $\mu$ and variance $\xi^2$ is
\begin{align}
\label{eq:g_pdf}
\mathsf{pdf}_{\mathcal{G}(\mu, \xi)}(x) = \frac{1}{\sqrt{2\pi}\xi} \exp\left(-\frac{(x-\mu)^2}{2 \xi^2}\right) 
\end{align} 

%: $f(D) + \mathcal{G}(\mathbf{0}, \Delta_2 f \cdot \sigma  )$, where 
%\begin{align}
%\label{eq:g_pdf}
%\mathsf{pdf}_{\mathcal{G}(\bm{\mu}, \Delta_2 f \cdot \sigma)}(\mathbf{x}) = \frac{\exp\left(-\frac{\sum_{i=1}^d(x_i-\mu_i)^2}{2 (\Delta_2 f \cdot \sigma)^2}\right)}{\sqrt{(2\pi)^d}(\sigma\Delta_2 f)^d} 
%\end{align} 
In fact, the Gaussian mechanism draws vector values from a multivariate spherical (or isotropic) Gaussian distribution
which is described by random variable $\mathcal{G}(f(D), \Delta_2 f \cdot \sigma\mathbf{I}_n)$, where $n$ is omitted if its unambiguous in the given context.

\subsection{Compressive Sensing}
\label{sec:cs}
%The Shannon-Nyquist theorem states that every continuous signal can be recovered from its discretization if its sampling rate is at least twice the maximum frequency present in the signal, which implicitly assumes that a signal is bandlimited. 

%The majority of the previous acquisition protocols are based on the Nyquist-Shannon sampling theorem.

Compressive Sensing (CS) introduced in \cite{donoho2006compressed,candes2006near,candes2006robust} aims to recover the original signal from significantly fewer samples (or measurements) than other traditional sampling techniques, which are based on  the Nyquist-Shannon theorem,  by exploiting the sparsity of the signal.    %compressibility.

Consider a signal $\mbf{x} \in \mathbb{R}^{n}$ which admits a sparse representation $\mbf{s} \in \mathbb{R}^{n}$, that is, there exists a \textit{sparsity orthonormal basis} with matrix $\Psi \in \mathbb{R}^{n \times n}$ such that: 
\begin{align}
\label{eq:cs_1}
\mbf{x}=\Psi \mbf{s} 
\end{align}
Here, $\mbf{s}$ is $U$-sparse if $\| \mbf{s} \|_{0}= \mbf{U}$. $\Psi$ can denote any linear transformation, such as Discrete Fourier/Cosine or Wavelet Transform, which render the original signal $\mbf{x}$ sparse. If $\mbf{x}$ is already sparse, then $\Psi$ can be the identity matrix which corresponds to the canonical sparsity basis.

In CS, $\mbf{x}$ is reconstructed from some of its \textit{linear measurements}.
For $m$ measurements, the signal is  ``sampled'' in $m$ values $\mbf{y}_j = \langle \phi_j, \mbf{x}\rangle$ $(1 \leq j \leq m)$, where the vectors $\phi_j \in \mathbb{R}^n$ constitute the sensing basis matrix $\Phi = (\phi_1, \phi_2, \ldots, \phi_m)^{\top} \in \mathbb{R}^{m\times n}$. Here, $m=r\times n$, where $r$ is the compression ratio. 
%The sampling is performed by using the \textit{sensing basis} matrix $\Phi \in \mathbb{R}^{m \times n}$. $\Phi=(\phi_{i,j})_{1\leq i,j \leq n}$ such that $\forall(i,j)\in [1,n]^{2}, i\neq j \Rightarrow \phi_{i,j}=0$, else: $\phi_{i,j}=1$.
Therefore, the compression operator $\mathcal{C}$ is defined as:
\begin{align}
\label{eq:cs_2}
\mathcal{C}(\mbf{x},m) =\mbf{y}= \Phi \mbf{x} = \Phi \Psi \mbf{s} = \Theta \mbf{s}
\end{align}
where $\Theta$ is the sparsity sensing matrix. 

There are several options to select the sensing matrix $\Phi$. When $\Phi$ is a random matrix (e.g., each element of $\Phi$ is an iid sample from   $\mathcal{G}(0, 1/m)$), then $\Psi$ works well with an arbitrary sparsity basis \cite{jacques2010compressed}. On the other hand, the numerical reconstruction of $\mbf{x}$ in that case has a complexity of $O(mn)$ which can be very large (recall that $n$ is the model size in the order of $10^6$). Another (faster) option for the sensing matrix $\Phi$ is when it is composed of random $m$ rows of the matrix of the (real) Discrete Fourier/Cosine Transform. Then, matrix multiplication can be executed with the Fast Fourier Transform (FFT) in $O(n\ln n)$, but such sensing matrix provides accurate reconstruction if $\Psi$ is the identity matrix, i.e. $\mbf{x}$ is already sparse \cite{jacques2010compressed}. Fortunately, this  usually holds for gradient vectors (or can be made as such by sparsification without significantly affecting convergence) and hence we will use this option in this paper.

In order to recover $\mbf{s}$ from $\mbf{y}$, one has to solve a system of linear equations with $m$ equations and $n$ unknowns. Although this system seems underdetermined because $m < n$, CS exploits the $U$-sparsity of $\mbf{s}$ for the reconstruction. 
%and we could then write an $m \times U$ system of linear equations with $m\geq U$.
It aims to reconstruct the sparse vector $\mbf{s}$ from $\mbf{y}= \Theta \mbf{s}$ given the sparsity sensing basis $\Theta$ by solving the following optimization problem: 
%and it may be solved by using a non-linear recovery optimizer:
\begin{align*}
    %\label{eq:cs_4}
   \arg \underset{s}{\min} \| \mbf{s} \|_{0} \quad  \textrm{s.t.} \quad \mbf{y}=\Theta \mbf{s}
\end{align*}

Since this optimization problem  is NP-complete \cite{natarajan1995sparse, jacques2010compressed}, it is further relaxed into the following slightly different problem called Basis Pursuit (BP) \cite{chen2001atomic}:
\begin{align*}
    %\label{eq:cs_5}
    \arg \underset{s}{\min} \| \mbf{s} \|_{1} \quad  \textrm{s.t.} \quad \mbf{y}=\Theta \mbf{s}
\end{align*}
Indeed, the convex $L_1$-norm usually approximates  the non-convex $L_0$-norm well, and the relaxed optimization problem can be efficiently solved with any convex optimization technique  \cite{jacques2010compressed} (e.g., with an LP solver).

%Hence, the problem was transformed to a much simpler convex problem, which can be easily solved by using standard optimization techniques .

When the measurements $\mbf{y}$ are noisy (i.e., $\mbf{y}=\Theta \mbf{s} + \mbf{z}$, where $\mbf{z}\in \mathbb{R}^{m}$  is the additional bounded iid noise, i.e. $\|\mbf{z}\|_{2}\leq \kappa$), then the following convex quadratic variant of BP called Basis Pursuit Denoising (BPDN) is rather considered: 
%A convex quadratic variant of BP called Basis Pursuit Denoising (BPDN) was defined as:
\begin{align*}
  %  \label{eq:cs_6}
   \mathcal{R}(\mathbf{y},\kappa) = \arg \underset{s}{\min} \| \mbf{s} \|_{1} \quad  \textrm{s.t.} \quad \|\mbf{y}-\Theta \mbf{s} \|_{2} \leq \kappa
\end{align*}
%When $\kappa = 0$, BPDN becomes BP. 
and therefore  
%constrained formulation of 
%The constrained formulation \ref{eq:cs_6} is equivalent to:
%\begin{align}
%    \label{eq:cs_7}
%    \mathcal{D}(y,n) = \arg \underset{s}{\min} \|y-\Theta s \|_{2} + \lambda \| s \|_{1},
%\end{align}
%We can rewrite \ref{eq:cs_7} by squaring the %first term and then make it differentiable:
\begin{align}
    \label{eq:cs_8}
    \mathcal{D}(\mbf{y},n) = \Psi\left(\arg \underset{s}{\min} \frac{1}{2}\|\mbf{y}-\Theta \mbf{s} \|_{2}^{2} + \lambda \| \mbf{s} \|_{1}\right),
\end{align}
Eq.~\eqref{eq:cs_8} defines our decompression operator and is an instance of convex quadratic programming. In this paper, we use the Orthant-Wise Limited-memory Quasi-Newton (OWL-QN) algorithm \cite{andrew2007scalable,pylbfgs}, an extension of Limited-memory BFGS, which is a numerical scalable optimization procedure  that can efficiently solve Eq.~\eqref{eq:cs_8}.

When $\lambda\rightarrow 0$, the problem in Eq.~\eqref{eq:cs_8} becomes  BP because $\lambda \| \mbf{s} \|_{1}$ tends to 0. 
In the case of non-noisy sensing measurements, a BP decoder is more adapted to reconstruct the sparse signal $s$. Otherwise, BPDN is more suited. This has particular importance in our case when the compressed vector (measurements) are noised to guarantee Differential Privacy, i.e.,  $\mbf{y}=\Theta \mbf{s} + \mbf{z}$  where $\mbf{z} \sim \mathcal{N}(0, S \mathbf{I}\sigma)$ (see Section \ref{sec:operation}).
Approximate signal reconstruction from noisy measurements have been theoretically justified in \cite{candes2008cs} from the Restricted Isometry Property of $\Theta$.

\begin{definition}[Restricted Isometry Property (RIP) \cite{candes2005decoding}]
The $U$-restricted isometry constants $0 \leq \delta_{U} < 1$ of a matrix $\Theta \in \mathbf{R}^{m \times n}$ is defined as the smallest number such that: 
\begin{align*}
%\label{eq:RIP}
(1-\delta_{U}) \| \mbf{s} \|_{2}^{2} \leq \| \Theta \mbf{s} \|_{2}^{2} \leq (1+\delta_{U}) \| \mbf{s} \|_{2}^{2}
\end{align*}
for all $U$-sparse vector $\mbf{s} \in \mathbb{R}^{n}$ and we say that the matrix $\Theta$ obeys the \textit{Restricted Isometry Property} (or RIP($U$,$\delta_{U}$)) of order $U < m$.
\label{def:RIP}
\end{definition}

\begin{theorem}[Reconstruction error of BPDN \cite{candes2008cs}]\label{thm:rip}
If $\Theta$ is $\text{RIP}(2U, \delta_U)$ and $\delta_U < \sqrt{2} - 1$, then 
$||\mathbf{s} - \mathcal{R}(\mathbf{y}, \kappa)||_2 \leq C \kappa + (D/\sqrt{K}) ||\mathbf{s} - \mathbf{s}_K ||_1$, where $C$ and $D$ are constants and $\mathbf{s}_K$ is a vector with all but the $K$-largest entries of $\mbf{s}$ set to zero\footnote{For instance, for $\delta_U = 0.2$, $C < 4.2$ and $D < 8.5$}.
\end{theorem}

%However, if $\lambda\rightarrow + \infty$, $s$ tends to 0 and then $\Theta s \rightarrow 0$,  $\lambda \| s \|_{1} \rightarrow y$.

%In the case of a non-noisy sensing measurements of the form $y=\Theta s$ the BP decoder is more adapted to reconstruct the sparse signal. However, if the measurement vector is noisy and has the form: $y=\Theta s + z$, where $z\in \mathbb{R}^{m}$  is the additional noise and it is assumed bounded, i.e. $\|z\|_{2}\leq \kappa$, then BPDN is more suited \cite{jacques2010compressed,chen2001atomic}. Surprisingly, an optimization algorithm which can solve a BPDN problem seems to be able also to reconstruct a signal based on noisy measurements, with a noise added by a differential private mechanism, where $z \sim \mathcal{N}(0, S \mathbf{I}\sigma)$.

%For our optimization problem we have decided to use the  Orthant-Wise Limited-memory Quasi-Newton (owlqn) algorithm \cite{andrew2007scalable}, a numerical scalable optimization procedure  that can efficiently find the optimum of an objective of the form: $f(s)= g(s) + \lambda \| s \|_{1}$, where $g(s)$ is a smooth function as in (\ref{eq:cs_8}).

Finally, notice that the compression operator $\mathcal{C}$ in Eq.~\eqref{eq:cs_2} is \emph{linear}, which means that:
\begin{align*}
    \sum_i \mathcal{C}(\mbf{x}_i,m) = \mathcal{C}\left(\sum_i \mbf{x}_i,m\right)
\end{align*}
and therefore
\begin{align*}
    \mathcal{D}\left(\sum_i \mathcal{C}(\mbf{x}_i,m)\right) \approx \sum_i \mbf{x}_i
\end{align*}
This linearity allows to combine secure aggregation and compressive sensing described in Section \ref{sec:operation}. 

\subsection{Error Propagation}
\label{sec:error}
Biased estimation of the gradients may prevent model convergence unless the approximation error introduced by lossy compression techniques, such as compressive sensing, sketching, or quantization, is accumulated and re-injected in every optimization round  \cite{Karimireddy} as follows:

\begin{align*}
    \mbf{g}_{t}=\nabla f(\mathbb{B}, \mbf{w}_{t-1}) &: \text{Computing gradients on batch $\mathbb{B}$} \\
    \mbf{p}_{t}=\eta \mbf{g}_{t} + \mbf{e}_{t-1} &: \text{Error feedback (correction)} \\
    \Delta_{t}= \mathcal{D}(\mathcal{C}(\mbf{p}_{t})) &: \text{Reconstruction of $\mbf{p}_t$} \\
    \mbf{w}_{t}= \mbf{w}_{t-1} -\Delta_{t} &: \text{Updating model parameters (weights)} \\
    \mbf{e}_{t}=\mbf{p}_{t}-\Delta_{t} &: \text{Error accumulation} 
\end{align*}

%At each iteration $t$, $\mbf{e}_{t}$ is the difference between the actual gradient $\mbf{p}$ and the new compressed gradient $\mbf{\Delta}$.

The corrected direction $\mbf{p}_{t}$ is obtained by adding the error $\mbf{e}_{t-1}$ accumulated over all iterations to $\mbf{g}_{t}$ (see Alg.~2 in \cite{Karimireddy} for more details). 
Here, the error is calculated based on the biased estimation of the update given by $\mathcal{D}(\mathcal{C}(\mbf{p}_t))$.

%The corrected direction $\mbf{p}_{t}$ is obtained by adding the error $\mbf{e}_{t-1}$ accumulated over all iterations to $\eta\mbf{g}_{t}$ (see Alg.~2 in \cite{Karimireddy} for more details). 

%Here, the error is calculated based on the biased estimation of the update given by $\mathcal{C}$, however, in our case, the biased estimation is instead given by $\mathcal{U}(\mathcal{C}(\cdot,m),n)$ and as it is computed on the server side then we can accumulate the error on it within a compressed vector $\mbf{e}_{t}$ of size $m$ instead of an update vector of size $n$. This solution was proposed in \cite{rothchild2020fetchsgd}. They also propose to maintain in addition a momentum vector of size m on the server side.

%Moreover, in \cite{rothchild2020fetchsgd} a solution was proposed to maintain the error accumulation vector at the server instead of at each client. Indeed, the nature of the federated learning and the low participation of each client during the learning process \cite{kairouz2019advances} does not allow to manage an error accumulation vector per client as it needs in this case the participation of all the clients during all the rounds. And because the bias is introduced at the server during the reconstruction step rather than during the compression at the client side, then the error can be accumulated by the server. They also include the momentum as it is also allowed by the compression operator. Indeed, error accumulation and momentum are linear operations and then needs a compression operator which is linear and supports linear operations.

\section{Federated Learning with Compressive Sensing}

\label{sec:fl_cs}

In the FL-STD scheme, presented in Section~\ref{sec:backg}, each randomly selected client sends its complete model update to the server. Knowing that a model has on average millions of parameters (each is a floating point value represented on 32 bits), the network can suffer from large traffic.

To decrease large network traffic, we adapt compressive sensing to federated learning. The new algoritm is called FL-CS. Moreover, this scheme is also extended to a privacy-preserving version, called FL-CS-DP, which aims to protect the training data of every participant. We show that compression improves model performance with Differential Privacy by reducing the added noise compared to the uncompressed DP variant of federated learning. Hence, both FL-CS and FL-CS-DP improve bandwidth efficiency, and in addition, FL-CS-DP also boosts the accuracy of differentially private federated learning. 

In what follows, we will first describe the non-private scheme FL-CS and then the privacy-preserving FL-CS-DP. 

%More specifically, FL-CS (see Alg.~\ref{alg:fed_cs}) differs from the standard federated scheme StdFed (see Alg.~\ref{alg:fed_learn}) as follows:

%\begin{enumerate}
%    \item Each client returns a proportion of coefficients from the frequency domain $y_{t}^{k}=\Theta(\mbf{w}_{t}^k- \mbf{w}_{t-1}^k)$ instead of sending the complete update in the time domain $\Delta \mbf{w}_{t}^{k}=(\mbf{w}_{t}^k- \mbf{w}_{t-1}^k)$. We use the discrete cosine transform (DCT) \cite{ahmed1974discrete,Ahmed91} to transform the weights vector from the time domain to the frequency domain.
%    \item The server averages the coefficient vectors  $y_{t}^{k}$ sent by each client $k$. The momentum and the error feedback are calculated before to reconstruct the vector $s_{t}$ the error is then accumulated for the next round. However, the error due to the compression was not considered in FL-Std (as we do not use compression for this scheme).
%    \item The FL-CS uses then the reconstructed vector $s_{t}$ to update the model: $\mbf{w}_{t} = \mbf{w}_{t-1} + \mbf{s}_{t} $, where, in FL-Std we update the model just after the averaging step: $\mbf{w}_{t} = \mbf{w}_{t-1} + \sum_{k} \frac{|D_k|}{\sum_j^N |D_j|} \Delta \mbf{w}_{t}^{k}$
%\end{enumerate}

\subsection{FL-CS: Federated Learning with  Compressive Sensing}

CS assumes the sparsity of the reconstructed signal in a specific basis domain $\Psi$ as explained in Section~\ref{sec:backg}. We assume the model update (as a signal) to be already sparse in the time domain, that is, $\Psi$ is canonical sparsity basis (i.e., $\Psi = \mathbf{I}$), and therefore, the compression operator is $\mathcal{C}(\Delta \mathbf{w}, m) = \Phi \Delta \mathbf{w}$, 
where $\Phi$ is composed of the first $m$ rows of the matrix of the Discrete Cosine Transform (DCT) \cite{ahmed1974discrete,Ahmed91}. Indeed, due to the large energy compaction property of DCT, the first coefficients, which correspond to the low frequency components of $\Delta \mathbf{w}$, tend to have the largest magnitude and hence convey the most information about the model update \cite{rao2014discrete}. In fact, for a canonical sparsity basis $\Psi = I$, $\Theta = \Phi$ is RIP with overwhelming probability as soon as $m = O(U \ln^4 n)$ if $\Delta \mathbf{w}$ is $U$-sparse \cite{tao06}. Therefore, reconstruction is possible according to Theorem \ref{thm:rip}.

The decompression operator $\mathcal{D}$ is defined Eq.~\eqref{eq:cs_8}. Note that the compression operator can be computed in $O(n \ln n)$ with FFT and the decompression (or reconstruction) operator is implemented with the OWL-QN algorithm \cite{andrew2007scalable} which makes our approach reasonably fast in practice.

%Note, also that the transformation from the time domain to the frequency domain is performed by the discrete cosine transform function and the inverse transformation by the inverse discrete cosine transform function. 

FL-CS is described in Alg.~\ref{alg:fed_cs}. A client first  computes its update $\Delta\mbf{w}_{t}^{k}$ with SGD, and then transfers the compressed update $\mathcal{C}(\Delta\mbf{w}_{t}^{k}, m)$, which consists of the first $m$ DCT coefficients of the update (Line 18). The server takes the average of the client's updates (Line 8), updates the momentum (Line 9), and computes the error $\mbf{e}_{t}$ (Line 10-12) due to compression following the error propagation technique described in Section \ref{sec:error}. This error is accumulated over all federated rounds and added to the model (Line 13) to compensate its negative effect on convergence. The server uses  OWL-QN \cite{andrew2007scalable,pylbfgs} to reconstruct the error-compensated aggregated model update $\mbf{s}_{t}\in \mathbb{R}^{n}$.
Finally, the server updates the global model as $\mbf{w}_{t} = \mbf{w}_{t-1} + \mbf{s}_{t}$ before re-distributing the updated model to a new set of clients $\mathbb{K}$.

Notice that the error $\mathbf{e}_t$, the averaged model update $\mathbf{y}_t$, as well as the momentum are maintained in the compressed domain and have a size of $m$ instead of $n$. This is possible due to the linearity of the compression scheme which is detailed in Section \ref{sec:cs}. \medskip

\textbf{Scalable reconstruction:}
Although OWL-QN is reasonably fast in practice, its computational overhead may not be tolerated with very large models.
A scalable reconstruction is proposed as follows.
On the client side, the update vector $\Delta \mathbf{w}_t$ is shuffled and then splitted into $P$ equally-sized chunks. Then, the compression operator $\mathcal{C}$ is applied on each individual chunk. Finally, the compressed chunks are transferred to the server.
On the server side, each chunk is reconstructed independently using OWL-QN.
The decompressed chunks are concatenated, and the resulted vector with size $n$ is reshuffled to obtain $\mathbf{s}_t$ by inverting the client-side shuffling.

Shuffling is performed by each client identically which guarantees that the compressed chunks can still be aggregated by the server. In practice, this can be implemented by sharing a common random seed among all participants to initialize the shuffler. As the server also knows this seed, it can invert this shuffling and reconstruct the aggregated model updates. 

Notice that, instead of reconstructing the complete update vector at once, the server performs reconstruction on smaller chunks which makes decompression faster. In addition, shuffling guarantees that the sparsity of the chunks is proportional to the sparsity of the whole update vector (i.e., if the update vector is $U$-sparse then all its chunks are $U/P$-sparse). Hence, the same compression operator $\mathcal{C}(\cdot, m/P)$ can be applied on every chunk without increasing the compression ratio (i.e., the compressed update still has a size of $m$). 

Note that shuffling is also performed identically over all the rounds to maintain the error.

%represented in the time domain basis $\Psi$ such that: $\hat{y_{t}}=\Phi\Psi\mbf{s}_{t}$ \footnote{$\hat{y_{t}}$ is represented in the frequency domain basis $\Psi^{-1}$ such that: $\mbf{s}_{t}=\Psi^{-1}\Phi^{-1}\hat{y_{t}}$}. 
%\todo{Move the next to the experimental section}
%As in \cite{ivkin2019communication}, we fix the momentum parameter $\rho$ to 0.9 and we tuned the global learning rate $\eta_{G}$. $\eta_{G}$ from 0.05 to 2.0 with an increment value of 0.05 and $P =200$.

%\subsection{Practical Training Algorithm}
%In practice, we modify FL-CS and FL-CS-DP in the following ways:

%\begin{itemize}
%    \item On the client side, we first shuffle the signal to uniform the sparsity and then we divide the signal into $P$ chunks ($P=200$ in our case), each of size $\frac{n}{P}$. After that we transform the signal from the time domain to the frequency domain and $\mathcal{C}$ to sample $r\times \frac{n}{P}$ coefficients per chunk.
%    \item On the server side, we will reconstruct each chunk independently instead of reconstructing the complete signal at once immediately. 
%\end{itemize}

%Dividing the signal on $P$ chunks will result on a scalable algorithm which can handle and reconstruct a large model whatever the power computations: on the client side for both the transformation and compression, on the server side for the reconstruction.

\subsection{FL-CS-DP: Differentially Private Federated Learning with Compressive Sensing}

\subsubsection{Privacy Model}
\label{sec:privacy_model}
We consider an adversary, or a set of colluding adversaries, who can access any update vector sent by the server or any clients at each round of the protocol.  A plausible adversary is a participating entity, i.e. a malicious client or server, that wants to infer the training data used by other participants.
The adversary is \emph{passive} (i.e., honest-but-curious), that is, it follows the learning protocol faithfully. 

Different privacy requirements can be considered depending on what information the adversary aims to infer. In general, private information can be inferred about:
\begin{itemize}
    \item any record (user) in any dataset of any client (\emph{record-level privacy}),
    \item any client/party (\emph{client-level privacy}).
\end{itemize}

To illustrate the above requirements, suppose that several banks build a common model to predict the creditworthiness of their customers. A bank certainly does not want other banks to learn the financial status of any of their customers (record privacy) and perhaps not even the average income of all their customers  (client privacy).
%\todo{another example for client privacy? One good reason can be: for the marketing strategy, if we know that the average income of all their customers is high, then the other banks may plan to open new offices in this location.}

Record-level privacy is a standard requirement used in the privacy literature and is usually weaker than client-level privacy. Indeed, client-level privacy requires to hide any information which is unique to a client including perhaps all its training data.   

%The motivation of client-based privacy is that several parties may contribute confidential data to build a common model, and they do not want their competitors (i.e., other parties or the aggregator) to learn \emph{any} information that is specific only to their training data.
%However, this requirement can be too strict in some settings and may require significant distortion of the training. Hence, a more relaxed requirement is to prevent the leakage of any information that is specific only to a single record or group of records in the training data.     
%For example, suppose that several banks build a common model to predict the creditworthiness of their customers. 
%A bank certainly does not want other banks to learn the financial status of any of their customers (record privacy), the number of their afro-american customers (group privacy) and perhaps neither any aggregate statistic about their customers such as their average income (client privacy).  

We aim at developing a solution that provides \emph{client-level privacy and is also bandwidth efficient}. 
For example, in the scenario of collaborating banks, we aim at protecting any information that is unique to each single bank's training data.
The adversary should not be able to learn from the received model or its updates whether any client's data is involved in the federated run (up to $\varepsilon$ and $\delta$). 
We believe that this adversarial model is reasonable in many practical applications when the confidential information spans over multiple samples in the training data of a single client (e.g., the presence of a group a samples, such as people from a certain race). Differential Privacy guarantees plausible deniability not only to any groups of samples of a client but also to any client in the federated run. Therefore, any negative privacy impact on a party (or its training samples) cannot be attributed to their involvement in the protocol run.

\subsubsection{Operation}
\label{sec:operation}

FL-CS-DP is described in Alg.~\ref{alg:fed_cs_dp}.
Client-level differential privacy requires each client to add Gaussian noise to the compressed model updates. In particular, each client first calculates $\mbf{c}_{t}^{k} = \mathcal{C}(\Delta \mathbf{w}_t^k, m)$ (in Line 19), which is then clipped (in Line 20) to obtain $\hat{\mbf{c}}_{t}^{k}$ with $L_2$-norm at most $S$. Then, random noise $\mbf{z}_{k} \sim \mathcal{G}(0,S\mathbb{\sigma}\mathbf{I}/\sqrt{\mathbb{K}})$ is added to $\hat{\mbf{c}}_{t}^{k}$ such that $\sum_{k \in \mathbb{K}} (\hat{\mbf{c}}_{t}^{k} +  \mbf{z}_{k}) = \sum_{k \in \mathbb{K}} \hat{\mbf{c}}_{t}^{k} + \mathcal{G}(0,S\mathbb{\sigma}\mathbf{I})$ as the sum of Gaussian random variables also follows Gaussian distribution\footnote{More precisely, $\sum_{i}\mathcal{G}(\nu_{i},\xi_{i})=\mathcal{G}(\sum_{i} \nu_{i},\sqrt{\sum_{i}\xi_{i}^{2}})$} and then differential privacy is satisfied where $\varepsilon$ and $\delta$ can be computed using the moments accountant described in Section \ref{sec:DP}.

However, as the noise is inversely proportional to $\sqrt{\mathbb{K}}$, 
$\mbf{z}_{k}$ is likely to be small if $|\mathbb{K}|$ is too large. Therefore, the adversary accessing an individual update $\hat{\mbf{c}}_{t}^{k} + \mbf{z}_{k}$ can almost learn a non-noisy update since $\mbf{z}_{k}$ is small. Hence, each client uses secure aggregation to encrypt its individual update before sending it to the server. Upon reception, the server sums the encrypted updates as:
\begin{align}
\sum_{k\in\mathbb{K}} \mathbf{y}_t^k &= \sum_{k\in\mathbb{K}}\mathsf{Enc}_{K_k}(\hat{\mbf{c}}_{t}^{k} + \mbf{z}_{k}) \nonumber \\
& = \sum_{k\in\mathbb{K}} \hat{\mbf{c}}_{t}^{k} + \sum_{k\in\mathbb{K}}\mbf{z}_{k} \nonumber \\
& = \sum_{k\in\mathbb{K}} \hat{\mbf{c}}_{t}^{k} + \mathcal{G}(0,S\mathbb{\sigma}\mathbf{I}) \label{eq:a1}
\end{align}
where $\mathsf{Enc}_{K_k}(\hat{\mbf{c}}_{t}^{k} + \mbf{z}_{k})= \hat{\mbf{c}}_{t}^{k} + \mbf{z}_{k} + \mbf{K}_k \mod p$ and $\sum_{k}\mbf{K}_k=0$ (see \cite{AcsC11,BonawitzIKMMPRS16} for details). Here the modulo is taken element-wise and $p=2^{\lceil \log_{2}(\max_{k}|| \hat{\mathbf{c}}^k_t + \mbf{z}_{k}||_{\infty}|\mathbb{K}|)\rceil}$.
Let $\gamma_t^k = 1/\max\left(1, \frac{||\mbf{c}_{t}^{k}||_2}{S}\right)$. Then, 
\begin{align}
\sum_{k \in \mathbb{K}} \hat{\mathbf{c}}_t^k &= \sum_{k\in \mathbb{K}} \gamma_t^k \mathbf{c}_t^k  \nonumber  \\
& = \sum_{k\in \mathbb{K}} \gamma_t^k \mathcal{C}(\Delta \mathbf{w}_t^k, m) \nonumber \\
& = \mathcal{C}(\sum_{k\in \mathbb{K}} \gamma_t^k \Delta \mathbf{w}_t^k, m) \label{eq:a2}
\end{align}
where the last equality comes from the linearity of the compression operation (see Section \ref{sec:cs}).
Plugging Eq.~\eqref{eq:a2} into Eq.~\eqref{eq:a1}. we get that
\begin{align*}
\sum_{k\in\mathbb{K}} \mathbf{y}_t^k = \mathcal{C}(\sum_{k\in \mathbb{K}} \gamma_t^k \Delta \mathbf{w}_t^k, m) + \mathcal{G}(0,S\mathbb{\sigma}\mathbf{I})
\end{align*}

This is an instance of BPDN (see Section \ref{sec:cs}), and therefore the direct reconstruction of $\sum_{k\in\mathbb{K}} \mathbf{y}_t^k$ would be an approximation of $\sum_{k\in \mathbb{K}} \gamma_t^k \Delta \mathbf{w}_t^k$. However, analogously to FL-CS, the server applies error propagation and computes the (noisy) error $\mathbf{e}_t$  from $\mathbf{y}_t = (1/|\mathbb{K}|) \sum_{k\in\mathbb{K}} \mathbf{y}_t^k$ (in Line 10), and decompresses $\mathbf{e}_t$ into $\mathbf{s}_t$ by using OWL-QN. Recall that the reconstruction algorithm solves the BPDN problem, where a sparse vector $\mathbf{s}$ is reconstructed from $m$ noisy measurements of the form $\Theta \mathbf{s} + \mathbf{z}$, where the noise $\mbf{z}\in\mathbb{R}^{m}$ is assumed to be identically and independently distributed over its elements with a Gaussian distribution \cite{jacques2010compressed,chen2001atomic}. 
Since $z \sim \mathcal{G}(0, S \mathbf{I}\sigma)$ in our case, the reconstruction algorithm is therefore optimized to reconstruct the differentially private compressed vectors (see Theorem \ref{thm:rip}).
\medskip

\noindent \textbf{Privacy analysis:}
The server can only access the noisy aggregate  which is sufficiently perturbed to ensure differential privacy; any client-specific information that could be inferred from the noisy aggregate is tracked and quantified by the moments accountant, described in Section~\ref{sec:DP}, as follows. 

Let $\eta_0(x|\xi) =  \mathsf{pdf}_{\mathcal{G}(0, \xi)}(x)$ and $\eta_1(x|\xi) =  (1-C) \mathsf{pdf}_{\mathcal{G}(0, \xi)}(x) + C \mathsf{pdf}_{\mathcal{G}(1, \xi)}(x)$ where $C$ is the sampling probability of a single client in a single round. Let
\begin{align}
\label{eq:alpha}
\alpha(\lambda| C) &= \log\max(E_1(\lambda, \xi, C), E_2(\lambda, \xi, C)) 
\end{align}
where
$%\begin{align*}
E_1(\lambda,  \xi, C) =  \int_{\mathbb{R}}\eta_0(x|\xi, C) \cdot \left(\frac{\eta_0(x|\xi, C)}{\eta_1(x|\xi, C)}\right)^{\lambda} dx
$ and
$ E_2(\lambda,  \xi, C) = \int_{\mathbb{R}}\eta_1(x|\xi, C) \cdot \left(\frac{\eta_1(x|\xi, C)}{\eta_0(x|\xi, C)}\right)^{\lambda} dx
$.

\begin{theorem}[Privacy of FL-CS-DP]
\label{thm:dg_privacy}
FL-CS-DP is $(\min_\lambda  (T_{\mathsf{cl}}\cdot \alpha (\lambda | C)  - \log \delta) /\lambda, \delta)$-DP. 
\end{theorem}
Given a fixed value of $\delta$, $\varepsilon$ is computed numerically  as in \cite{Abadi,MironovTZ19}.

The magnitude of the added Gaussian noise is proportional to the sensitivity $S$, which is in turn often proportional to the model size $n$ \cite{zhu2020votingbased}. 
Hence, when $n$ becomes large, SGD often fails to converge due to the perturbation error caused by the added noise  \cite{zhu2020votingbased}.
In our approach, the perturbation error is less since Gaussian noise is added to the compressed vector with size $m < n$.
On the other hand, compression also induces some reconstruction error owing to its lossy nature.
The total error is the sum of the reconstruction and the perturbation error and is quantified in Theorem \ref{thm:rip}. 
Finding the right trade-off between these two errors is the key to achieve good model quality.

\begin{algorithm}[h]
\small
		\caption{FL-CS: Federated Learning \label{alg:fed_cs}}
	\DontPrintSemicolon
	{\bf Server:}\;
	\Indp Initialize common model $w_0$ , $\eta_{G}$ , $\rho$, $\mbf{u}_{t}=0$, $\mbf{e}_{t}=0$ \, 
	
	\For {$t=1$ \KwTo $\Tcl$}
	{
	    Select $\mathbb{K}$ clients uniformly at random \;
		\For {\textrm{each} client $k$ \textrm{in} $\mathbb{K}$}
		{	
			$ \mbf{y}_t^k = \mathbf{Client}_k(\mbf{w}_{t-1})$\;
		}
		
		$\mbf{y}_{t}=\sum_{k=1}^{|\mathbb{K}|} \frac{ \mbf{y}_{t}^{k}}{|\mathbb{K}|}$ : Averaging
		
		$\mbf{u}_{t}=\rho\mbf{u}_{t-1}+\mbf{y}_{t}$ : Momentum
		
		$\mbf{e}_{t}=\eta_{G}\mbf{u}_{t}+\mbf{e}_{t-1}$ : Error Feedback
		
		$\mbf{s}_{t}= \mathcal{D}(\mbf{e}_{t},n)$     : Reconstruction
		
		%$\hat{\mbf{y}_{t}}=\Psi\mbf{s}_{t}$ : Transform to Frequency domain
		
		%$\mbf{e}_{t}=\mbf{e}_{t}-\mathcal{C}(\hat{\mbf{y}_{t}},\mbf{m})$ :  Error accumulation
		
		$\mbf{e}_{t}=\mbf{e}_{t}-\mathcal{C}(\mbf{s}_{t},m)$ :  Error accumulation

		$\mbf{w}_{t} = \mbf{w}_{t-1} + \mbf{s}_{t} $ : Update \;
	}
	\KwOut{Global model $\mbf{w}_t$}\;
	\Indm {\bf $\mathbf{Client}_{k}(\mbf{w}_{t-1}^k)$:}\;
	\Indp
	$\mbf{w}_{t}^k = \mathbf{SGD}(D_k, \mbf{w}_{t-1}^k, \Tgd)$\;
	$\Delta \mbf{w}_{t}^{k}=\mbf{w}_{t}^k- \mbf{w}_{t-1}^k$\;
	%$\hat{\Delta \mbf{w}_{t}^{k}}=\Psi\Delta \mbf{w}_{t}^{k}$: Transform to Frequency domain\;
	\KwOut{Model update $\mathcal{C}(\Delta \mbf{w}_{t}^{k},m)$} 
\end{algorithm}

\begin{algorithm}[h]
\small
		\caption{FL-CS-DP: Private Compressive Sensing Federated Learning \label{alg:fed_cs_dp}}
	\DontPrintSemicolon
	{\bf Server:}\;
	\Indp Initialize common model $w_0$ , $\eta_{G}$ , $\rho$, $\mbf{u}_{t}=0$, $\mbf{e}_{t}=0$ \, 
	
	\For {$t=1$ \KwTo $\Tcl$}
	{
	    Select $\mathbb{K}$ clients uniformly at random \;
		\For {\textrm{each} client $k$ \textrm{in} $\mathbb{K}$}
		{	
			$ \mbf{y}_t^k = \mathbf{Client}_k(\mbf{w}_{t-1})$\;
		}
		
		$\mbf{y}_{t}=\sum_{k=1}^{|\mathbb{K}|} \frac{ \mbf{y}_{t}^{k}}{|\mathbb{K}|}$ : Averaging
		
		$\mbf{u}_{t}=\rho\mbf{u}_{t-1}+\mbf{y}_{t}$ : Momentum
		
		$\mbf{e}_{t}=\eta_{G}\mbf{u}_{t}+\mbf{e}_{t-1}$ : Error Feedback
		
		$\mbf{s}_{t}= \mathcal{D}(\mbf{e}_{t},\mbf{n})$     : Reconstruction
		
		%$\hat{\mbf{y}_{t}}=\Psi\mbf{s}_{t}$ : Transform to Frequency domain
		
		$\mbf{e}_{t}=\mbf{e}_{t}-\mathcal{C}(\mbf{s}_{t},m)$ :  Error accumulation
		
		$\mbf{w}_{t} = \mbf{w}_{t-1} + \mbf{s}_{t} $ : Update \;
	}
	\KwOut{Global model $\mbf{w}_t$}\;
	\Indm {\bf $\mathbf{Client}_{k}(\mbf{w}_{t-1}^k)$:}\;
	\Indp
	$\mbf{w}_{t}^k = \mathbf{SGD}(D_k, \mbf{w}_{t-1}^k, \Tgd)$\;
	$\Delta \mbf{w}_{t}^{k}=\mbf{w}_{t}^k- \mbf{w}_{t-1}^k$\;
	%$\hat{\Delta \mbf{w}_{t}^{k}}=\Psi\Delta \mbf{w}_{t}^{k}$: Transform to Frequency domain\;
	
	$\mbf{c}_t^k = \mathcal{C}(\Delta \mbf{w}_{t}^{k},m)$\;
	
	$\hat{\mbf{c}}_t^k = \mbf{c}_{t}^{k} / \max\left(1, \frac{||\mbf{c}_{t}^{k}||_2}{S}\right)$\;
	
	\KwOut{$\mathsf{Enc}_{K_k}(\mathcal{G}(\hat{\mbf{c}}_t^k, S \mathbf{I}\sigma /\sqrt{|K|}))$} 
\end{algorithm}

\begin{algorithm}[h]
\small
		\caption{FL-STD-DP: Federated Learning with Client Privacy \label{alg:fl_std_dp}}
	\DontPrintSemicolon
	{\bf Server:}\;
	\Indp Initialize common model $w_0$\;
	\For {$t=1$ \KwTo $\Tcl$}
	{
	    Select $\mathbb{K}$ clients randomly \;
		\For {each client $k$ \textrm{in} $\mathbb{K}$}
		{	
			$\Delta \tilde{\mathbf{w}}_t^k = \mathbf{Client}_k(\mbf{w}_{t-1})$\;
		}
		$\mbf{w}_{t} = \mbf{w}_{t-1} + \frac{1}{|\mathbb{K}|} \sum_{k} \Delta \tilde{\mathbf{w}}_t^k $\;
	}
    \Indm {\bf $\mathbf{Client}_{k}(\mbf{w})$:}\;
    \Indp
	$\mbf{w}_{t-1}^k = \mbf{w}$\;
	$\Delta \mbf{w}_t^k = \mathbf{SGD}(D_k, \mbf{w}_t^{k-1}, \Tgd) - \mbf{w}_{t-1}^k$\;
	$\Delta \hat{\mbf{w}}_t^k = \Delta \mbf{w}_t^k / \max\left(1, \frac{||\Delta \mbf{w}_t^k||_2}{S}\right)$\;
    \KwOut{$\mathsf{Enc}_{K_k}(\mathcal{G}(\Delta \hat{\mbf{w}}_t^k, S \mathbf{I}\sigma /\sqrt{|K|}))$}
\end{algorithm}

\section{Experimental Results}
\label{sec:xp}

The goal of this section is to evaluate the performance of our proposed schemes FL-CS and FL-CS-DP on a benchmark dataset and a realistic in-hospital mortality prediction scenario. We aim at evaluating their performance with different levels of compression and comparing them with the performance of the  following learning protocols:
\begin{itemize}
    \item FL-STD: It is described in Section \ref{FL-STANDARD} (see Alg.~\ref{alg:fed_learn}). %The standard federated learning algorithm described in \ref{alg:fed_learn} consists of choosing randomly a subset of clients $\mathbb{K}$, each client update the received model on his local dataset before to send it back to the server. After that, the server has only to average and update the model.
    \item FL-RND: This baseline follows the  algorithm of FL-STD except that a random subset of the update vector with size  $m\leq n$ is sent to the server instead of the complete update of size $n$. Each client selects the same random subset of coordinates from the update vector, but a different subset in every round. The server then averages the received updates before updating only the corresponding $m$ weights. Note that if $m=n$, FL-RND is equivalent to FL-STD (see Alg.~\ref{alg:fl_rnd}).
    \item FL-FREQ: In this baseline, a client  transforms the model update to the frequency domain by using DCT \cite{ahmed1974discrete,Ahmed91}, and then the first $m$ coefficients (low frequency components) are extracted and sent to the server as in FL-CS. However, as opposed to FL-CS, the server reconstructs the aggregated update vector by applying the inverse DCT on the aggregated compressed vectors where the last $n-m$ coefficients are zeroed out  (see Alg.~\ref{alg:fl_freq}). This baseline corresponds to a low-pass filter applied on the update vector. $\Phi$ in Alg.~\ref{alg:fl_freq} is composed of the first $m$ rows of the matrix of the DCT.

\end{itemize}

\subsection{Medical Dataset}

\subsubsection{The In-hospital Mortality Prediction Scenario}

The ability to accurately predict the risks in the patient's perspectives of evolution is a crucial prerequisite in order to adapt the care that certain patients receive \cite{Papier_Amela}.

We consider the scenario where several hospitals are collaborating to train models for in-hospital mortality prediction using
our Federated Learning schemes. 
This well-studied real-world problem consists in trying to precisely identify the patients who are at risk of dying from complications during their hospital stay \cite{Avati2018,rajkomar-npj18,Papier_Amela}. As commonly found in the literature \cite{Papier_Amela}, for such predictions, we focus on hospital admissions of adults hospitalized for at least 3 days, excluding elective admissions.

%The main goal of this experimental section is to evaluate the performance of FL-SIGN-DP with different levels of privacy (i.e. different $\epsilon$ values) and compare it with the performance of the  following learning protocols:
%\begin{itemize}
%    \item \emph{(Non-federated) Centralized training}: The training data of all hospitals are merged %and a single model is trained on this merged data without any privacy guarantee.
%    \item \emph{FL-STANDARD} is described in Section \ref{FL-STANDARD}.
%    \item \emph{FL-SIGN} is described in  Section \ref{sec:fl-sign-desc}.
%    \item \emph{FL-STANDARD-DP}  is specified in Alg.~\ref{alg:basic_rec_fed_learn}.
%    It has the same privacy guarantee as FL-SIGN-DP\footnote{the privacy analysis in Section %\ref{sec:priv_anal} also applies to FL-STANDARD-DP} but is less bandwith efficient; its communication %overhead is identical to that of FL-STANDARD.
%    Specifically, unlike in FL-SIGN-DP, each client sends the original (non-quantized) update vector %$\mbf{s}_{t}^{k} = \mbf{w}_t^k - \mbf{w}_{t-1}^k$ to the server. Then, the server computes the model %update as $\mbf{w}_{t} = \mbf{w}_{t-1} + \frac{1}{\mathbb{|K|}}\left(\sum_{k}  %\mbf{s}_{t}^{k}\right)$. Apart from these differences, FL-STANDARD-DP is identical to FL-SIGN-DP.  
%\end{itemize}

\subsubsection{The Premier Healthcare Database}

We used EHR data from the Premier healthcare database\footnote{\href{https://www.premierinc.com/newsroom/education/premier-healthcare-database-whitepaper}{https://www.premierinc.com/newsroom/education/premier-healthcare-database-whitepaper}} which is one of the largest clinical databases in the United States, collecting information from millions of patients over a period of 12 months from 415 hospitals in the USA \cite{Papier_Amela}. These hospitals are supposedly representative of the United States hospital experience \cite{Papier_Amela}. Each hospital in the database provides discharge files that are dated records of all billable items (including therapeutic and diagnostic procedures, medication, and laboratory usage) which are all linked to a given patient's admission \cite{Papier_Amela,Makadia}. 

The initial snapshot of the database used in our work (before pre-processing step) comprises the EHR data of 1,271,733 hospital admissions.
Electronic Health Record (EHR) is a digital version of a patient’s paper chart readily available in hospitals. For developing supervised learning and specifically deep learning models, we focus on a specific set of features from EHR data. The features of interest that capture the patients information are summarized in Table~\ref{tab:data_description}. There is a total of 24,428 features per patient, mainly due to the variety of drugs possibly served. As in \cite{Avati2018}, we also removed all the features which appear on less than 100 patients' records, hence, the number of features was reduced to 7,280 features. 

The Medication regimen complexity index (MRCI) \cite{MRCI} is an aggregate score computed from a total of 65 items, whose purpose is to indicate the complexity of the patient's situation. The minimum MRCI score for a patient is 1.5, which represents a single tablet or capsule taken once a day as needed (single medication). However the maximum is not defined since the number of medications increases the score \cite{MRCI}. In our case, after statistical analysis of our dataset, we consider the MRCI score as ranging from 2 to 60.

\begin{table*}[!h]
	\caption{Descriptions of features} 
	\label{tab:data_description}
	\begin{tabular}{|r|l|}
		\hline
		     \emph{Features}    			            & \emph{Descriptions} 		                 \\
		\hline
		Age	                & 	 Value in the range of 15 and 89	  \\
		\hline
		Gender               &   Male, Female or Unknown  \\
		\hline
		Admission type     &   Emergency, Urgent, Trauma Center: visits to a trauma center/hospital or Unknown \\ %\tablefootnote{\url{https://www.resdac.org/cms-data/variables/claim-inpatient-admission-type-code-ffs}}  \\ 
		\hline
		MRCI               &   Medication regimen complexity index score (ranging from 2 to 60)   \\
		\hline
		Drugs               &  Drugs given to the patient on the $1^{st}$ day of hospitalization. There is a total of 24,419 possible drugs that can be served.  \\
		\hline
	\end{tabular}
\end{table*}

Most real datasets like ours are generally imbalanced with a skewed distribution between the classes. In our case, the positive cases (patients who die during their hospital stay) represent only 3\% of all patients. Table \ref{tab:class_prop} gives more details about this distribution after the pre-processing step which is discussed in \ref{preprocessing}. To deal with this well-known problem, we have decided to use downsampling technique \cite{more2016survey,he2009learning}, a standard solution used for this purpose \footnote{We have also tested weighted loss function and oversampling techniques. But, we noticed experimentally that downsampling technique outperforms the others whatever the considered scheme.}.

\subsubsection{Model architecture}

As in \cite{Avati2018}, we use a fully connected neural network model with the following architecture: two hidden layers of 200 units, which use a Relu activation function followed by an output layer of 1 unit with sigmoid activation function and a binary cross entropy loss function. A dropout layer with a rate set to 0.5 is used between each hidden layer and between the last hidden layer and the output layer. This results in 1,496,601 parameters in total. We tune $\eta$ from 0.01 to 0.5 with an increment value of 0.005. As in \cite{ivkin2019communication}, we fix the momentum parameter $\rho$ to 0.9 and we tuned the global learning rate $\eta_{G}$ from 0.05 to 2.0 with an increment value of 0.05. The number of chunks is $P =200$. The hyperparameters used by each of the considered schemes are summarized in Table~\ref{tab:S_table_medical}.

The sensitivity $S$ is selected during an initialization round for each scheme by taking the median value over $N$ $L_2$-norm values. We also noticed that the sensitivity of FL-CS-DP, FL-RND and FL-FREQ are nearly equivalent for the same level of compression. For this reason, the same sensitivity value is used for each compressed scheme and for the same compression ratio. Table~\ref{tab:S_table_fashion_mnist} and Table~\ref{tab:S_table_medical} show the selected clipping threshold (i.e., sensitivity $S$) for each dataset and according to each compression ratio.

\begin{table}[h!]
	\caption{Number of instances for our case study. The Medical dataset contains in total 1,271,733 records.}
	\label{tab:class_prop}
	\centering
	\begin{tabular}{ccccc}
		\hline
		Data & Positive cases         			                & Negative cases		                            & Ratio     & Total   \\
		\hline
		Train & 	      32,106  	                & 		     985,280                           &      3.16\%     &  1,017,386  \\
		\hline
		Test & 	        7,882	                & 		        246,465                        & 3.10\%          &  254,347 \\
		\hline
	\end{tabular}
\end{table}

\subsection{Fashion-MNIST}
\subsubsection{Data Description}
Fashion-MNIST database of fashion articles consists of 60,000 28x28 grayscale images of 10 fashion categories, along with a test set of 10,000 images \cite{Fashion-MNIST} \cite{KERAS_datasets}.
\subsubsection{Data pre-processing \& experimental setup} \medskip

\noindent \textbf{Preprocessing:}
The pixel of each image is an unsigned integer in the range between 0 and 255. We rescale them to the range [0,1] instead. \medskip

\noindent \textbf{Model architecture:}
For Fashion-MNIST, we use a model \cite{FedAVG} with the following architecture: a convolutional neural network (CNN) with two 5x5 convolution layers (the first with 32 filters, the second with 64, each followed with 2x2 max pooling), a fully connected layer with 512 units and ReLu activation, and a final softmax output layer. This results in 1,663,370 parameters in total. We tune $\eta$ from 0.01 to 0.5 with an increment value of 0.005. As in \cite{ivkin2019communication}, we fix the momentum parameter $\rho$ to 0.9 and we tuned the global learning rate $\eta_{G}$ from 0.05 to 2.0 with an increment value of 0.05. The number of chunks used is $P =200$. The hyperparameters used by each of the considered schemes are summarized in Table~\ref{tab:S_table_fashion_mnist}.

\subsection{Computational environment}

Our experiments were performed on a server running Ubuntu 18.04 LTS equipped with a Intel(R) Xeon(R) Silver 4114 CPU @ 2.20GHz, 192GB RAM, and two NVIDIA Quadro P5000 GPU card of 16 Go each. We use Keras 2.2.0 \cite{KERAS} with a TensorFlow  backend 1.12.0 \cite{TensorFlow} and Numpy 1.14.3 \cite{Numpy} to implement our models and experiments. We use  Python 3.6.5 and our code runs on a Docker container to simplify reproducibility.

\subsection{Results}

Table~\ref{tab:description_results_Fashion_MNIST} represents the best accuracy over 200 rounds for each scheme on the Fashion-Mnist dataset. $\mathit{Round}$ corresponds to the round when the best accuracy is reached and $\mathit{Cost}$ is the average bandwidth consumption calculated as: $r \times n \times 32 \times \mathit{Round} \times C$, where 32 is the number of bits necessary to represent a float value, $n$ is the uncompressed model size, $r=\frac{m}{n}$, $m$ is the compressed model size, $C$ is the sampling probability of a client, and $\mathit{Round}$ is the round when we get the the best accuracy.

Table~\ref{tab:description_results_Medical_data} represents the best balanced accuracy over 100 rounds for each scheme on the Medical dataset. AUROC (area under the receiver operating characteristic curve \cite{AUROC}) corresponds to the AUROC value when the best balanced accuracy is reached, round is also the round when we get the best balanced accuracy, and finally, Cost is the average bandwidth consumption calculated as for the Fashion-MNIST dataset described above.

Without DP, notice that our FL-CS scheme outperforms FL-RND and FL-FREQ whatever the considered compression ratio or the dataset are. Also, compared to FL-STD, our scheme started to reach the same accuracy from a compression ratio $r$ being equal or greater than 0.1 for both datasets, although the differences between FL-CS and FL-STD for a compression ratio of 0.05 are only of 6\%~\footnote{Based on the accuracy} and 1\%~\footnote{Based on the balanced accuracy \cite{Balanced_acc_1,Balanced_acc_2}} for the Fashion-Mnist and the medical datasets, respectively. However, FL-STD consumes much more bandwidth than FL-CS. Indeed, FL-CS reduces the bandwidth cost by 95\% compared to FL-STD with a compression ratio of 0.05 for both datasets, while the bandwidth cost is reduced to 80\% and 85\% with a compression ratio of 0.2 for Fashion-MNIST and the medical data, respectively.

Surprisingly, for the smallest compression ratio 5\%, FL-CS-DP performs as well as FL-RND-DP or FL-FREQ in terms of accuracy and much better in terms of bandwidth consumption. Indeed, FL-CS-DP with a compression ratio of 5\% reached 0.78 of accuracy on Fashion-MNIST, however, our baseline FL-RND needs a compression ratio of 10\% to reach a similar result (0.77) and 20\% to have slightly better result (0.80). The same holds for the medical dataset, where FL-CS-DP reached 0.69 and 0.76 of balanced accuracy and Auroc, respectively. However, our other baseline FL-FREQ needs a compression ratio of 20\% to reach the same performance. 
FL-CS-DP performs better for a small compression ratio. Indeed, FL-CS-DP reaches 0.78, 0.73 and 0.66 for 5\%, 10\% and 20\% of accuracy, respectively, on Fashion-MNIST. The accuracy degradation with DP can be explained by the fact increasing the compression ratio $r$ also increases the sensitivity $S$ which has a direct impact on the additive Gaussian noise as explained in Section~\ref{sec:operation}. Indeed, the standard deviation of the normal distribution is $\sigma \times S$.

On both datasets, FL-STD-DP suffers from the noise due to the large sensitivity which is the largest one in Table~\ref{tab:S_table_fashion_mnist} and \ref{tab:S_table_medical} for a compression ratio of 1.0 (uncompressed model). Even for  FL-RND and FL-FREQ, the gap between the non-private and the private version is larger when the compression ratio increases for both datasets. 
As the noise proportional to $S$ and is added to every coordinate, the norm of the added noise increases with the model size $n$. This has negative impact on model convergence for a large $n$ as discussed in \cite{zhu2020votingbased}. By decreasing $n$, compression helps reach better utility.

On Fashion-MNIST, FL-CS-DP with a compression ratio of 0.05 outperforms FL-STD-DP on both utility and bandwidth preservation. However, and even though FL-CS-DP reduces the bandwidth cost by 95\% which is not negligible, they have both comparable accuracy on the medical data. It can be explained by the reduction of the noise due to the reduction of $S$ and $\sigma$ (see Table~\ref{tab:S_table_medical} and Table~\ref{tab:S_table_fashion_mnist}) needed to reach an $\epsilon$ value of at most 1 after $\Tcl$ rounds. 

There is a possible tradeoff between the privacy, communication cost, and utility. Indeed, having a small $\epsilon$ (better privacy) results in a reduction of the communication costs while it decreases accuracy. FL-STD-DP, for example, converges to the best accuracy (61\%) after only 25 rounds with early stopping, which results on high privacy ($\epsilon$=0.69) and low communication cost (only 22.18 Megabyte). However, the accuracy degradation is more important (about 30\% which is the worst accuracy degradation indicated in Table~\ref{tab:description_results_Fashion_MNIST}). Indeed, the large amount of added noise impacts the convergence of the model which can not achieve an accuracy larger than 61\%.

Finally, we highlight a trade-off for  FL-CS-DP. As mentioned above, FL-CS-DP performs better when the smallest compression ratio $r$ is used, as the sensitivity for this level of compression is the smallest one. On the other hand, the compression ratio cannot be decreased arbitrarily as it will result in large reconstruction error. Therefore, one has to find the smallest compression ratio thay is small enough to reduce the perturbation error but large enough to induce small reconstruction error.

\begin{table*}[!ht]
    \centering
    \begin{tabular}{|c|c|c|c|c|c|}
        \hline
         \multirow{2}{*}{\emph{Compression ratio ($r$)}} & \multirow{2}{*}{Algorithms} & \multicolumn{4}{c|}{\emph{Performance}} \\
         \cline{3-6}
         %\hline
         & &  \emph{Accuracy}  & \emph{Round} & \emph{Cost (Megabyte)} & \emph{$\epsilon$} \\
         \hline

        \multirow{4}{*}{$0.05$} &  FL-RND  &  0.73  & 192 & 8.52 & N/A  \\
        \cline{2-6}
        &  FL-FREQ  &  0.73  & 189 & 8.38 & N/A  \\
        \cline{2-6}
        &  FL-CS   & 0.82  & 200 & 8,87  & N/A \\
        \cline{2-6}
        &  FL-RND-DP & 0.73  & 196 & 8.69  & 0.99\\ 
        \cline{2-6}
        &  FL-FREQ-DP  & 0.72  & 200 &  8,87 & 1 \\
        \cline{2-6}
        &  FL-CS-DP  & 0.78  & 197 &  8.74 & 1 \\
        \hline 
        \hline
        \multirow{4}{*}{$0.1$} &  FL-RND  & 0.78   & 200 & 17.74 & N/A  \\
        \cline{2-6}
        &  FL-FREQ  &  0.78  & 197 & 17.48 & N/A  \\
        \cline{2-6}
        &  FL-CS  & 0.85  & 199 &  17.65 & N/A \\
        \cline{2-6}
        %&  FL-CS + clip &   &  &   & N/A \\
        %\cline{2-6}
        &  FL-RND-DP  & 0.77  & 199 & 17.65  & 1\\
        \cline{2-6}
        &  FL-FREQ-DP  &  0.76 & 200 & 17.74  & 1 \\
        \cline{2-6}
        &  FL-CS-DP  & 0.73 & 101 & 8,96  & 0.84 \\
        \hline 
        \hline
        \multirow{4}{*}{$0.2$} &  FL-RND  & 0.82   & 200 & 35,49 & N/A  \\
        \cline{2-6}
        &  FL-FREQ &  0.82  & 195 & 34,60 & N/A  \\
        \cline{2-6}
        &  FL-CS  & 0.87  & 193 & 34,24  & N/A \\
        \cline{2-6}
        %&  FL-CS + clip &   &  &   & N/A \\
        %\cline{2-6}
        &  FL-RND-DP  &  0.80 & 199 & 35.31  & 1\\
        \cline{2-6}
        &  FL-FREQ-DP  & 0.79  & 200 & 35,49  & 1 \\
        \cline{2-6}
        &  FL-CS-DP  & 0.66  & 150 & 26,61  & 0.92 \\
        \hline 
        \hline

        \multirow{2}{*}{$1.0$} &  FL-STD  &  0.87  & 191 & 169.44 & N/A  \\
        \cline{2-6}
        &  FL-STD-DP  & 0.61  & 25 & 22.18  & 0.69 \\
        \hline
        
    \end{tabular}
    \caption{Summary of results on Fashion-MNIST dataset.}
    \label{tab:description_results_Fashion_MNIST}
\vspace{-.3cm}
\end{table*}

%%%%%%%%%%%%%%%%%%%%%%%%%%%%%%%%%%%%%%%%%%%%%%%%%%%%%%%%%%%%%%%%%%%%%%%%

\begin{table*}[!ht]
    \centering
    \begin{tabular}{|c|c|c|c|c|c|c|}
        \hline
         \multirow{2}{*}{\emph{Compression ratio ($r$)}} & \multirow{2}{*}{Algorithms} & \multicolumn{5}{c|}{\emph{Performance}} \\
         \cline{3-7}
         %\hline
         & &   \emph{Bal\_Acc} & \emph{AUROC}  & \emph{Round} & \emph{Cost(Megabyte)} & \emph{$\epsilon$} \\
         \hline

        \multirow{4}{*}{$0.05$} &  FL-RND  & 0.60 & 0.69 & 99 & 4.73 &  N/A \\
        \cline{2-7}
        &  FL-FREQ  & 0.69 & 0.76 & 100 & 4.78 & N/A  \\
        \cline{2-7}
        &  FL-CS &   0.73 & 0.80 & 100 & 4.78 & N/A \\
        %\cline{2-7}
        %&  FL-CS (First coefficient) &  0.66  & 0.72 & 100 & 4.78 & N/A \\
        %\cline{2-7}
        %&  FL-CS + (50\% First  +50 \%random ) &  0.71    & 0.78 & 98 & 4.68 & N/A \\
        \cline{2-7}
        &  FL-RND-DP &   0.60 & 0.69 & 100 & 4.78 & 1 \\
        \cline{2-7}
        &  FL-FREQ-DP  &   0.65  & 0.72 & 100 & 4.78 & 1 \\
        \cline{2-7}
        &  FL-CS-DP &  0.69 & 0.76 & 100 & 4.78 & 1\\
        %\cline{2-7}
        %&  FL-CS-DP (First coefficient) & 0.59  & 0.69 & 100 & 4.78 & 1\\
        %\cline{2-7}
        %&  FL-CS-DP (50\% First  +50 \%random )  &  0.62   & 0.68 & 87 & 4.16 & 0.97 \\
        \hline 
        \hline
        \multirow{4}{*}{$0.1$} &  FL-RND  & 0.66 & 0.73 & 100 & 9.56 &  N/A \\
        \cline{2-7}
        &  FL-FREQ &  0.71  & 0.78 & 100 & 9.56 & N/A  \\
        \cline{2-7}
        &  FL-CS  & 0.73 & 0.81 & 87 & 8.31 & N/A \\
        %\cline{2-7}
        %&  FL-CS (First coefficient)  & 0.69 & 0.76 & 100 & 9.56 & N/A \\
        %\cline{2-7}
        %&  FL-CS (50\% First  +50 \%random ) &  0.72  & 0.79 & 89 & 8.51 & N/A \\
        \cline{2-7}
        &  FL-RND-DP  & 0.65 & 0.72 & 100 & 9.56 & 1 \\
        \cline{2-7}
        &  FL-FREQ-DP  & 0.67  & 0.74 & 100 & 9.56 & 1 \\
        \cline{2-7}
        &  FL-CS-DP & 0.69 & 0.76 & 99 & 9.46 & 1\\
        %\cline{2-7}
        %&  FL-CS-DP  (First coefficient) & 0.64 & 0.71 & 100 & 9.56 & 1 \\
        %\cline{2-7}
        %&  FL-CS-DP (50\% First  +50 \%random ) &  0.64  & 0.71 & 99 & 9.46 & 1\\
        \hline 
        \hline
        
        \multirow{4}{*}{$0.2$} &  FL-RND   & 0.69 & 0.76 & 100 & 19.11 &  N/A  \\
        \cline{2-7}
        &  FL-FREQ & 0.72   & 0.80 & 100  & 19.11 & N/A  \\
        \cline{2-7}
        &  FL-CS  & 0.73 & 0.81 & 74 & 14.14 & N/A \\
        %\cline{2-7}
        %&  FL-CS (First coefficient)  & 0.72 & 0.79 & 99 & 18.92 & N/A \\
        %\cline{2-7}
        %&  FL-CS (50\% First  +50 \%random ) & 0.73  & 0.80 & 97 & 18.54 & N/A \\
        \cline{2-7}
        &  FL-RND-DP & 0.67 & 0.74 & 99 & 18.92 & 1 \\
        \cline{2-7}
        &  FL-FREQ-DP &  0.69  & 0.76 & 100 & 19.11 & 1\\
        \cline{2-7}
        &  FL-CS-DP & 0.68 & 0.74 & 64 & 12.23 & 0.92 \\
        %\cline{2-7}
        %&  FL-CS-DP (First coefficient) & 0.67 & 0.73 & 98 & 18.73 & 1 \\
        %\cline{2-7}
        %&  FL-CS-DP (50\% First  +50 \%random ) & 0.67  & 0.73 & 99 & 18.92 & 1\\
        \hline 
        \hline

        \multirow{2}{*}{$1.0$} &  FL-STD  & 0.74 & 0.82 & 99 & 94.62 & N/A \\
        \cline{2-7}
        &  FL-STD-DP   & 0.70 & 0.77 & 93 & 88.88 & 0.99 \\
        %\cline{2-9}
        %&  FL-STANDARD-DP + clip &    &  &  & & & & \\
        \hline 
        
    \end{tabular}
    \caption{Summary of results on Medical dataset.}
    \label{tab:description_results_Medical_data}
\vspace{-.3cm}
\end{table*}

\section{Related work}
\label{sec:related_work}
%CS technique was used for two purpose in general: compression or denoising. It was also used for two types of data: signals or images. 

\noindent \textbf{Privacy of Federated Learning:}
There exist a few inference attacks specifically designed against federated learning schemes. In \cite{Property_inference}, the adversary's goal is to infer whether records with a specific property are included in the training dataset of the other participants (called batch property inference). The authors demonstrate the attack by inferring whether black people are included in any of the training datasets, where the common model is trained for gender classification (i.e., the inferred property is independent of the learning objective). The adversary is supposed to have access to the aggregated model update of honest participants.
In \cite{NasrSH19}, the proposed attack infers if a specific person is included in the training dataset of the participants (aka, Membership inference). The adversary extracts the following features from every snapshot of the common model, which is a neural network: output value, hidden layers, loss values, and the gradient of the loss with respect to the parameters of each layer. These features are used to train a membership inference model, which is a convolutional neural network.

The concept of Client-based Differential Privacy has been introduced in \cite{Client-DP-McMahan} and \cite{Client-DP-ETH-Zurich}, where the goal is to hide any information that is specific to a single client's training data. These algorithms bound and noise the contribution of a single client's instead of a single record in the client's dataset. The noise is added by the server, hence, unlike our solution, these works assume that the server is trusted. Also, the noise is drawn from continuous distributions. \smallskip

\noindent \textbf{Bandwidth Optimization in Federated Learning:}
Different quantization methods have been proposed to save the bandwidth and reduce the communication costs in federated learning. They can be divided into two main groups: unbiased and biased methods. The unbiased approximation techniques use probabilistic quantization schemes to compress the stochastic gradient and attempt to approximate the true gradient value as much as possible \cite{QSGD}\cite{TERNGRAD}\cite{ATOMO}\cite{Quant_Fed}. 
However, biased approximations of the stochastic gradient can still guarantee convergence both in theory and practice \cite{SIGNSGD, LinHM0D18, SeideFDLY14}. In signSGD \cite{SIGNSGD}, all the clients calculate the stochastic gradient based on a single mini-batch and then send the sign vector of this gradient to the server. The server calculates the aggregated sign vector by taking the median (majority vote) and sends the signs of the aggregated signs back to each client.

A different line of works exploit the sparsity of model updates to compress model updates. Our work belongs to this line.
The authors in \cite{mamaghanian2011compressed} use CS for low-complexity energy-efficient ECG compression. 
Although compressed sensing was primarily designed for compression  \cite{candes2006robust,donoho2006compressed}, it was extended for denoising as in \cite{metzler2016denoising,tavakoli2012image} where compressive sensing is used for the purpose of denoising. 
In \cite{yu2014compressed}, compressed sensing based denoising and certain artificial intelligence are combined to improve the prediction performance.

CS was also used with DP in \cite{DP_CS}. The authors show that the amount of noise is reduced from $O(\sqrt{n})$ to $O(\log(n))$, when the noise is added on the sampled coefficients instead of the original database. 

Existing works \cite{mohamedamiri,mohamedamiri2}  proposed to use a compressive sensing for federated learning in order to compress model updates without privacy guarantees. However, they assume that all clients participate in each round (as they maintain an error accumulation vector at each client due to the compression scheme), but as discussed in \cite{kairouz2019advances} this assumption is not always realistic. Recently in \cite{jeon2020compressive} another compressive sensing algorithm was proposed for federated learning  for the denoising purpose (instead of the compression), where the added noise is due to the network transmission.  

Sketching was adapted to federated learning for the purpose of compressing model updates in \cite{ivkin2019communication}. The authors proposed to use Count-Sketch \cite{charikar2002finding} to retrieve the largest weights in the update vector on the server side. After that, the server uses two additional communication rounds to inform the clients about what gradient values they need to send back to the server. The server then takes the average of the received gradients and zeros-out the others before updating the model. The error due to the compression is maintained at each client, and the participation of all clients are required in each round which, as per \cite{kairouz2019advances} and as discussed above, is not practical to federated learning. In \cite{rothchild2020fetchsgd}, the aforementioned scheme is improved further by directly retrieving the most updated gradient values without asking for their positions in the update vector. This  makes the scheme more efficient as it needs fewer communication rounds. %remove the two additional round used to send the coefficients of the most updated weights from the server to the clients and the values of those weights from the clients to the server. 
Similarly to our approach, the error vector is also maintained on the server side instead of the client side, which is clearly a better fit for federated learning.

%One major challenge in system design is to reconstruct local gradient vectors accurately at the central server, which are computed-and-sent from the wireless devices. To overcome this challenge, we first establish a transmission strategy to construct sparse transmitted signals from the local gradient vectors at the devices. We then propose a compressive sensing algorithm enabling the server to iteratively find the linear minimum-mean-square-error (LMMSE) estimate of the transmitted signal by exploiting its sparsity.

\section{Conclusion}

%We propose a scalable compressive sensing solution designed for Federated learning to reduce the communication cost. 

In this paper, we propose to extend Federated Learning with compressive sensing.
Specifically, we propose two schemes: the first one (FL-CS) uses compressive sensing in order to reduce communication bandwidth. The second one (FL-CS-DP) 
combines compressive sensing and differential privacy in order to protect participants' information.
 
We present some experimental results that are based on the Fashion-MNIST dataset as well as on a  medical dataset of 1.2 millions of US hospital patients. 
Results indicate that using compressive sensing in Federated Learning allows to reduce the communication costs by up to 95\% for a moderate loss of accuracy.

Results with the privacy-preserving extension FL-CS-DP indicate that compression happens to be especially useful and interesting in this context as it improves accuracy.
This is due to the sensitivity reduction which is proportional to the added noise needed to guarantee differential privacy, and to the considered optimization problem (BPDN) which reconstructs data from noisy measurements.  

We believe that the proposed privacy-preserving extension (FL-CS-DP) is an interesting alternative to differentially private federated learning (FL-STD-DP), as it improves both accuracy and bandwidth cost.

\begin{table*}[]
    \centering
    \begin{tabular}{|c|c|}
        \hline
          \emph{Algos}     &     \emph{Parameters}  \\
         \cline{1-2}
         %\hline
        \multirow{3}{*}{FL-STD \& FL-STD-DP (r=1.0)}   & $S=2.15$; $C=1/60$; $N=6000$; $\Tcl=200$; $\Tgd=5$; $|\mathbb{B}|=10$; \\ & $|D_k|=10$; $n=1,663,370$; $\delta=10^{-5}$; $SGD(\eta=0.215)$; $\sigma=1.54$ \\
        \cline{1-2}
        \multirow{3}{*}{FL-CS,FL-RND,FL-FREQ and their private extensions (r=0.2)}  & $S=0.98$; $C=1/60$; $N=6000$; $\Tcl=200$; $\Tgd=5$; $|\mathbb{B}|=10$; \\ & $|D_k|=10$; $n=1,663,370$; $\delta=10^{-5}$; $SGD(\eta=0.215)$; \\ & $\eta_{G}=0.35$; $\rho=0.9$; $P=200$; $\sigma=1.54$  \\
        \cline{1-2}
        \multirow{3}{*}{FL-CS,FL-RND,FL-FREQ and their private extensions (r=0.1)}  &  $S=0.69$; $C=1/60$; $N=6000$; $\Tcl=200$; $\Tgd=5$; $|\mathbb{B}|=10$; \\ & $|D_k|=10$; $n=1,663,370$; $\delta=10^{-5}$; $SGD(\eta=0.215)$; \\ & $\eta_{G}=0.35$; $\rho=0.9$; $P=200$; $\sigma=1.54$  \\
        \cline{1-2}
         \multirow{3}{*}{FL-CS,FL-RND,FL-FREQ and their private extensions (r=0.05)}  & $S=0.47$; $C=1/60$; $N=6000$; $\Tcl=200$; $\Tgd=5$; $|\mathbb{B}|=10$; \\ & $|D_k|=10$; $n=1,663,370$; $\delta=10^{-5}$; $SGD(\eta=0.215)$; \\ & $\eta_{G}=0.35$; $\rho=0.9$; $P=200$; $\sigma=1.54$  \\
        \cline{1-2}
        \hline 
        
    \end{tabular}
    \caption{Common environment between the schemes on Fashion-MNIST. $\rho$, $\eta_{G}$ and $P$ are only used with FL-CS and FL-CS-DP.}
    \label{tab:S_table_fashion_mnist}
\vspace{-.3cm}
\end{table*}

\begin{table*}[]
    \centering
    \begin{tabular}{|c|c|}
        \hline
          \emph{Algos}     &     \emph{Parameters}  \\
         \cline{1-2}
         %\hline
        \multirow{2}{*}{FL-STD \& FL-STD-DP (r=1.0)}   & $S=0.31$; $C=100/5011$; $N=5011$; $\Tcl=100$; $\Tgd=5$; \\ &  $n=1,496,601$; $\delta=10^{-5}$; $SGD(\eta=0.1)$ ; $\sigma=1.49$\\
        \cline{1-2}
        \multirow{2}{*}{FL-CS,FL-RND,FL-FREQ and their private extensions (r=0.2)}  & $S=0.14$; $C=100/5011$; $N=5011$; $\Tcl=100$; $\Tgd=5$; \\ &  $n=1,496,601$; $\delta=10^{-5}$; $SGD(\eta=0.1)$; $\eta_{G}=1.0$; $\rho=0.9$; $P=200$; $\sigma=1.49$  \\
        \cline{1-2}
        \multirow{2}{*}{FL-CS,FL-RND,FL-FREQ and their private extensions (r=0.1)}  &  $S=0.1$; $C=100/5011$; $N=5011$; $\Tcl=100$; $\Tgd=5$; \\ &  $n=1,496,601$; $\delta=10^{-5}$; $SGD(\eta=0.1)$; $\eta_{G}=1.0$; $\rho=0.9$; $P=200$; $\sigma=1.49$  \\
        \cline{1-2}
         \multirow{2}{*}{FL-CS,FL-RND,FL-FREQ and their private extensions (r=0.05)}  & $S=0.07$; $C=100/5011$; $N=5011$; $\Tcl=100$; $\Tgd=5$; \\ & $n=1,496,601$; $\delta=10^{-5}$; $SGD(\eta=0.1)$; $\eta_{G}=1.0$; $\rho=0.9$; $P=200$; $\sigma=1.49$  \\
        \cline{1-2}
        \hline 
        
    \end{tabular}
    \caption{Common environment between the schemes on the Medical dataset. $\rho$, $\eta_{G}$ and $P$ are only used with FL-CS and FL-CS-DP.}
    \label{tab:S_table_medical}
\vspace{-.3cm}
\end{table*}

\bibliographystyle{IEEEbib}
\bibliography{references}

\appendices

\section{Medical data: Data pre-processing \& experimental setup details}

This section describes the experimental setting which is used to evaluate the accuracy and the privacy of our proposals.

\subsection{Preprocessing} 

\label{preprocessing}

\begin{enumerate}
    \item \textbf{Features normalization}: we extract from the dataset the values of each feature represented in Table \ref{tab:data_description}. For gender, we use one-hot encoding: Male, Female and Unknown. Similarly, for admission type we use 4 features: Emergency, Urgent, Trauma Center, and Unknown \footnote{\url{https://www.resdac.org/cms-data/variables/claim-inpatient-admission-type-code-ffs}}. For drugs, we extract 24,419 features which correspond to the different drugs (name and dosage). A given patient receives only a few of the possible drugs served, resulting in a very sparse patient's record. We use a MinMax normalization for age and MRCI in order to rescale the values of these features between 0 and 1 (using MinMaxScaler class of scikit-learn\footnote{\url{https://scikit-learn.org/stable/modules/generated/sklearn.preprocessing.MinMaxScaler.html}}). The labels that we consider are boolean: true means that the patient died during his hospital stay while false means she survived.

    \item \textbf{Patients filtering}: We consider patient and drug information of the first day at the hospital so that we can make predictions 24 hours after admission (as commonly found in the literature \cite{rajkomar-npj18,Papier_Amela}). We filter out the pregnant and new-born patients because the medication types and admission services are not the same for theses two categories of patients. Our model prediction is built without patients' historical medical data. This has the advantage to require minimum patient's information and to work for new patients.
    
    \item \textbf{Hospitals filtering}: The dataset contains 415 hospitals for a total size of 1,271,733 records. We split randomly the dataset into disjoint training and testing data (80\% and 20\% respectively).  The final dataset for testing contains 254,347 patients, with 7,882 deceased patients and 246,465 non-deceased patients (see Table~\ref{tab:class_prop}).
    
    Using Client-Level differential privacy requires to add more noise than Record-Level differential privacy, because the privacy purposes are not the same as detailled in Section~\ref{sec:backg}. To reduce the noise (when $\epsilon$ is fixed) and then improve the utility, we have to reduce the number of iterations or to reduce the sampling probability which are the parameters used to compute $\epsilon$. We therefore have two options to reduce the sampling probability:
    \begin{itemize}
        \item[-] Reducing the number of clients selected at each round $|\mathbb{K}|$. However this option also decreases the amount of data, and hence have a negative impact on the utility. We therefore preferred to use the next option.
        \item[-] Increasing the total number of clients $N$: we created more hospitals by splitting randomly the training data over 5011 "virtual" hospitals. We also, took care to have at least one in-hospital dead patient per hospital. Each hospital contains 203 patients except one which has 356 patients. We created 5011 hospitals in order to have approximately the same number of patients per hospital, each of them with some in-hospital dead patients. 
        
        In practise, Client-Level differential privacy is more adapted to an environment with a large set of clients as explained in \cite{Client-DP-McMahan,Client-DP-ETH-Zurich}.
    \end{itemize}

\end{enumerate}

\subsection{Imbalanced data}

The dataset of each hospital is imbalanced because the proportion of patients that leave the hospital alive is, fortunately, much larger than in-hospital dead patients. To deal with this well-known problem, we have decided to use downsampling technique \cite{more2016survey,he2009learning}, a standard solution used for this purpose. \footnote{We have also tested weighted loss function and oversampling techniques. But, we noticed experimentally that downsampling technique outperforms the other techniques for all the schemes.}

%\subsubsection{Selection of hyperparameters}
%\todo{Remove?}
%The selection of hyperparameters in FL-SIGN-DP and FL-STANDARD-DP, such as batch size $|\mathbb{B}|$, scaling factor $\gamma$,  or sensitivity $S$, should also be differentially private. In order to compute them, one option is to use public datasets that have similar distribution as the clients' private training data. Another option is to use more sophisticated methods like the one exposed in Appendix D of \cite{Abadi}. 

%The hyperparameters in Table~\ref{tab:fl_sign_param_privacy} are tuned, and we select those which provide the best results for each scheme.

\subsection{Performance Metrics}

We use the following metrics:
\begin{itemize}
    \item \emph{Balanced accuracy}  \cite{Balanced_acc_1,Balanced_acc_2} is computed as  $1/2 \cdot (\frac{\TP}{\Pos}+\frac{\TN}{\Neg}) = \frac{\TPR+\TNR}{2}$ and is mainly used with imbalanced data. \emph{True Positive Rate} ($\TPR$) and \emph{True Negative Rate} ($\TNR$):
    $\TPR=\frac{\TP}{\Pos}$ and $\TNR=\frac{\TN}{\Neg}$, where $\Pos$ and $\Neg$ are the number of positive and negative instances, respectively, and $\TP$ and $\TN$ are the number of true positive and true negative instances.
    We note that traditional (``non-balanced'') accuracy metrics such as $\frac{\TP+\TN}{\Pos+\Neg}$ can be misleading for very imbalanced data \cite{akosa2017predictive}: in our dataset, the minority class has only 3\% of all the training samples (see Table~\ref{tab:class_prop}), which means that a biased (and totally useless) model always predicting the majority class would have a (non-balanced) accuracy of  $97\%$. 
    \item The \emph{area under the ROC curve} (\AUC) is also a frequently used accuracy metric. The ROC curve is calculated by varying the prediction threshold from 1 to 0, when \TPR and \FPR are calculated at each threshold. The area under this curve is then used to measure the quality of the predictions. A random guess has an $\AUC$ value of 0.5, whereas a perfect prediction has the largest $\AUC$ value of 1.  
\end{itemize}

\subsection{Evaluation Method.}
First, we split randomly the dataset of each hospital into disjoint training and testing data (80\% and 20\% respectively).
An entire federated run is executed with this split, and all the metrics are evaluated in every round on the union of all clients' testing data. 
All metric values of the round with the best balanced metric are recorded.

\begin{algorithm}[h]
\small
		\caption{FL-RND \label{alg:fl_rnd}}
	\DontPrintSemicolon
	{\bf Server:}\;
	\Indp Initialize common model $w_0$\;
	\For {$t=1$ \KwTo $\Tcl$}
	{
	    Generate a random seed $\zeta$
	    Select $\mathbb{K}$ clients uniformly at random \;
		\For {\textrm{each} client $k$ \textrm{in} $\mathbb{K}$}
		{	
			$\mbf{y}_t^k = \mathbf{Client}_k(\mbf{w}_{t-1},\zeta)$\;
		}

		$ \mbf{y}_{t}=\sum_{k} \frac{|D_k|}{\sum_j^N |D_j|} \Delta \mbf{w}_{t}^{k}$\;
		$j=0$\;
		\For {\textrm{each} element $i$ \textrm{in} $\mbf{G}$}
	    {
		    $\mbf{w}_{t}[i] = \mbf{w}_{t-1}[i] + \mbf{y}_{t}[j] $\;
	        $j=j+1$
	    }
	}
	\KwOut{Global model $\mbf{w}_t$}\;
	\Indm {\bf $\mathbf{Client}_{k}(\mbf{w}_{t-1}^k,\zeta)$:}\;
	\Indp
	$\mbf{w}_{t}^k = \mathbf{SGD}(D_k, \mbf{w}_{t-1}^k, \Tgd)$\;
	$\Delta \mbf{w}_{t}^{k}=\mbf{w}_{t}^k- \mbf{w}_{t-1}^k$ \;
	Generates a random set $\mbf{G}=\{x \in \{1, \cdots,n\}\}$ of $\mbf{m}$ random integer values such that $\mbf{m}\leq \mbf{n}$ based on the seed $\zeta$\;
	
	$\hat{\Delta w_{t}^{k}}=$Sample $\mbf{m}$ elements from $\Delta \mbf{w}_{t}^{k}$ by taking each element of $\mbf{G}$ as a coordinate \;
	
	\KwOut{The sampled Model update $\hat{\Delta w_{t}^{k}}$} 
	
\end{algorithm}

\begin{algorithm}[h]
\small
		\caption{FL-FREQ \label{alg:fl_freq}}
	\DontPrintSemicolon
	{\bf Server:}\;
	\Indp Initialize common model $w_0$\;
	\For {$t=1$ \KwTo $\Tcl$}
	{
	    Select $\mathbb{K}$ clients uniformly at random \;
		\For {\textrm{each} client $k$ \textrm{in} $\mathbb{K}$}
		{	
			$\Delta \mbf{y}_t^k = \mathbf{Client}_k(\mbf{w}_{t-1})$\;
		}

		$ \mbf{y}_{t}=\sum_{k} \frac{|D_k|}{\sum_j^N |D_j|} \Delta \mbf{w}_{t}^{k}$
		
		%$ \hat{\mbf{y}_{t}}=\Psi^{-1}\Phi^{-1}  \mbf{y}_{t}$
		
		$ \hat{\mbf{y}_{t}}=\Phi^{-1}  \mbf{y}_{t}$ : Transform to time domain \;
		
		$\mbf{w}_{t} = \mbf{w}_{t-1} +  \hat{\mbf{y}_{t}}$ \;
	}
	\KwOut{Global model $\mbf{w}_t$}\;
	\Indm {\bf $\mathbf{Client}_{k}(\mbf{w}_{t-1}^k)$:}\;
	\Indp
	$\mbf{w}_{t}^k = \mathbf{SGD}(D_k, \mbf{w}_{t-1}^k, \Tgd)$\;
	$\Delta \mbf{w}_{t}^{k}=\mbf{w}_{t}^k- \mbf{w}_{t-1}^k$\;
	%$\hat{\Delta \mbf{w}_{t}^{k}}=\Psi\Delta \mbf{w}_{t}^{k}$: Transform to Frequency domain\;
	%\KwOut{The sampled Model update $\mathcal{C}(\hat{\Delta \mbf{w}_{t}^{k}},m)$ } 
	
	\KwOut{The sampled Model update $\Phi \Delta \mbf{w}_{t}^{k}$ } 
	
\end{algorithm}

\begin{algorithm}[h]
\small
		\caption{FL-RND-DP \label{alg:fl_rnd_dp}}
	\DontPrintSemicolon
	{\bf Server:}\;
	\Indp Initialize common model $w_0$\;
	\For {$t=1$ \KwTo $\Tcl$}
	{
	    Generate a random seed $\zeta$
	    Select $\mathbb{K}$ clients uniformly at random \;
		\For {\textrm{each} client $k$ \textrm{in} $\mathbb{K}$}
		{	
			$\mbf{y}_t^k = \mathbf{Client}_k(\mbf{w}_{t-1},\zeta)$\;
		}

		$ \mbf{y}_{t}=\sum_{k} \frac{|D_k|}{\sum_j^N |D_j|} \Delta \mbf{w}_{t}^{k}$\;
		$j=0$\;
		\For {\textrm{each} element $i$ \textrm{in} $\mbf{G}$}
	    {
		    $\mbf{w}_{t}[i] = \mbf{w}_{t-1}[i] + \mbf{y}_{t}[j] $\;
	        $j=j+1$
	    }
	}
	\KwOut{Global model $\mbf{w}_t$}\;
	\Indm {\bf $\mathbf{Client}_{k}(\mbf{w}_{t-1}^k,\zeta)$:}\;
	\Indp
	$\mbf{w}_{t}^k = \mathbf{SGD}(D_k, \mbf{w}_{t-1}^k, \Tgd)$\;
	$\Delta \mbf{w}_{t}^{k}=\mbf{w}_{t}^k- \mbf{w}_{t-1}^k$ \;
	Generates a random set $\mbf{G}=\{x \in \{1, \cdots,n\}\}$ of $\mbf{m}$ random integer values such that $\mbf{m}\leq \mbf{n}$ based on the seed $\zeta$\;
	
	$\hat{\Delta w_{t}^{k}}=$Sample $\mbf{m}$ elements from $\Delta \mbf{w}_{t}^{k}$ by taking each element of $\mbf{G}$ as a coordinate \;
	
	$\hat{\Delta w_{t}^{k}}' = \hat{\Delta w_{t}^{k}} / \max\left(1, \frac{||\hat{\Delta w_{t}^{k}}||_2}{S}\right)$\;
    \KwOut{$\mathsf{Enc}_{K_k}(\mathcal{G}(\hat{\Delta w_{t}^{k}}', S \mathbf{I}\sigma /\sqrt{|K|}))$}
	
\end{algorithm}

\begin{algorithm}[h]
\small
		\caption{FL-FREQ-DP \label{alg:fl_freq_dp}}
	\DontPrintSemicolon
	{\bf Server:}\;
	\Indp Initialize common model $w_0$\;
	\For {$t=1$ \KwTo $\Tcl$}
	{
	    Select $\mathbb{K}$ clients uniformly at random \;
		\For {\textrm{each} client $k$ \textrm{in} $\mathbb{K}$}
		{	
			$\Delta \mbf{y}_t^k = \mathbf{Client}_k(\mbf{w}_{t-1})$\;
		}

		$ \mbf{y}_{t}=\sum_{k} \frac{|D_k|}{\sum_j^N |D_j|} \Delta \mbf{w}_{t}^{k}$
		
		$ \hat{\mbf{y}_{t}}=\Phi^{-1}  \mbf{y}_{t}$ : Transform to time domain \;
		
		$\mbf{w}_{t} = \mbf{w}_{t-1} +  \hat{\mbf{y}_{t}}$ \;
	}
	\KwOut{Global model $\mbf{w}_t$}\;
	\Indm {\bf $\mathbf{Client}_{k}(\mbf{w}_{t-1}^k)$:}\;
	\Indp
	$\mbf{w}_{t}^k = \mathbf{SGD}(D_k, \mbf{w}_{t-1}^k, \Tgd)$\;
	$\Delta \mbf{w}_{t}^{k}=\mbf{w}_{t}^k- \mbf{w}_{t-1}^k$\;
	$\hat{\Delta w_{t}^{k}} = \Phi \Delta \mbf{w}_{t}^{k} / \max\left(1, \frac{||\mathcal{C}(\hat{\Delta \mbf{w}_{t}^{k}},m)||_2}{S}\right)$\;
    \KwOut{$\mathsf{Enc}_{K_k}(\mathcal{G}(\hat{\Delta w_{t}^{k}}, S \mathbf{I}\sigma /\sqrt{|K|}))$}

\end{algorithm}

\end{document}